\documentclass[Afour,sageh,times]{sagej}

\usepackage{moreverb,url}

\usepackage[colorlinks,bookmarksopen,bookmarksnumbered,citecolor=red,urlcolor=red]{hyperref}

\usepackage{enumitem}
\usepackage{amsmath,amsfonts}
\usepackage{algorithmic}
\usepackage{algorithm}
\usepackage{array}
\usepackage[caption=false,font=normalsize,labelfont=sf,textfont=sf]{subfig}
\usepackage{textcomp}
\usepackage{stfloats}
\usepackage{verbatim}
\usepackage{graphicx}
\usepackage{cite}
\usepackage{graphics} 
\usepackage{epsfig}
\usepackage{times}
\usepackage{amssymb}
\usepackage{float}
\usepackage{arydshln}
\usepackage{multirow}
\usepackage{diagbox}
\usepackage{bbm}
\usepackage{makecell}
\usepackage{lscape}
\usepackage{mathbbol}
\usepackage{caption}

\newcommand{\etal}{\textit{et~al}.}
\newcommand{\ie}{\textit{i}.\textit{e}.}
\newcommand{\eg}{\textit{e}.\textit{g}.}

\newcommand\BibTeX{{\rmfamily B\kern-.05em \textsc{i\kern-.025em b}\kern-.08em
T\kern-.1667em\lower.7ex\hbox{E}\kern-.125emX}}

\def\volumeyear{2024}
\begin{document}

\runninghead{Pan \etal: DM-OSVP++}

\title{DM-OSVP++: One-Shot View Planning Using 3D Diffusion Models for Active RGB-Based Object Reconstruction}

\author{Sicong Pan\affilnum{1}\affilnum{2}\affilnum{3}, Liren Jin\affilnum{1}\affilnum{2}, Xuying Huang\affilnum{1}\affilnum{2}\affilnum{3}, Cyrill Stachniss\affilnum{1}\affilnum{2}\affilnum{3}, Marija Popovi\'{c}\affilnum{4} and Maren Bennewitz\affilnum{1}\affilnum{2}\affilnum{3}}

\affiliation{\affilnum{1} University of Bonn, Germany\\
\affilnum{2} Center for Robotics, Bonn, Germany\\
\affilnum{3} Lamarr Institute for Machine Learning and Artificial Intelligence, Germany\\
\affilnum{4} Delft University of Technology, The Netherlands
}

\corrauth{Sicong Pan, Humanoid Robots Lab, University of Bonn, Germany}

\email{span@uni-bonn.de}

\begin{abstract}
Active object reconstruction is crucial for many robotic applications. A key aspect in these scenarios is generating object-specific view configurations to obtain informative measurements for reconstruction. One-shot view planning enables efficient data collection by predicting all views at once, eliminating the need for time-consuming online replanning. Our primary insight is to leverage the generative power of 3D diffusion models as valuable prior information. By conditioning on initial multi-view images, we exploit the priors from the 3D diffusion model to generate an approximate object model, serving as the foundation for our view planning. Our novel approach integrates the geometric and textural distributions of the object model into the view planning process, generating views that focus on the complex parts of the object to be reconstructed. We validate the proposed active object reconstruction system through both simulation and real-world experiments, demonstrating the effectiveness of using 3D diffusion priors for one-shot view planning.
\end{abstract}


\maketitle

\section{Introduction} \label{sec:Introduction}

\begin{figure*}[!t]
\centering
\includegraphics[width=1.0\textwidth]{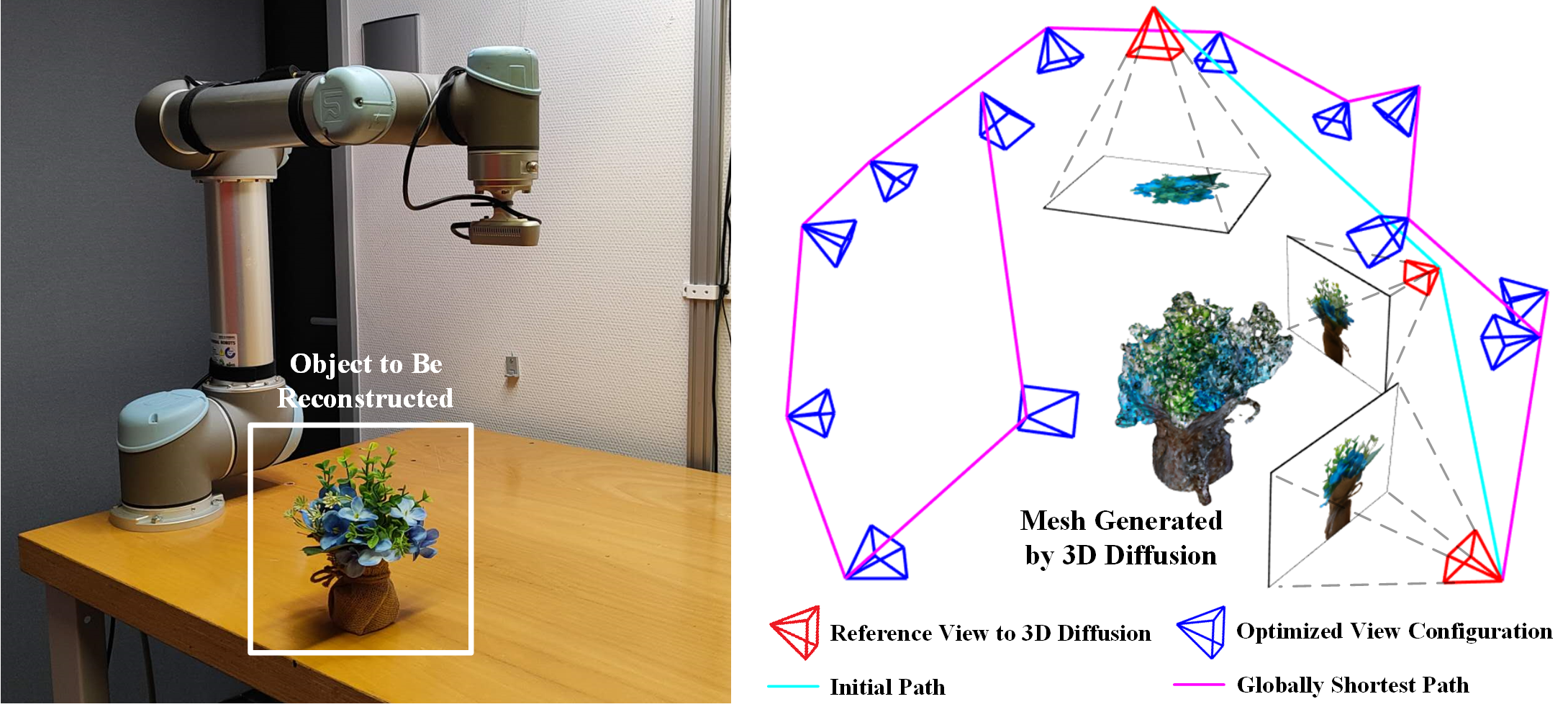}
\caption{
An example of our one-shot view planning in a real-world scenario.
Our goal is to actively plan a set of views (blue) at once to collect informative RGB images for object reconstruction using the robotic arm. 
The key component in our approach is a multi-image-to-3D diffusion model generating the corresponding 3D mesh by conditioning on several RGB images captured from the initial reference views (red).
By leveraging the mesh as both geometric and textural priors, our approach outputs view configurations specifically associated with the target object and calculates the globally shortest path.
In particular, our method plans denser views to observe more complex parts (the upper part of the object in the example) to improve the reconstruction quality.
} 
\label{fig_paper_cover}
\end{figure*}

Many autonomous robotic applications depend on accurate 3D models of objects to perform downstream tasks.
These include object manipulation in household scenarios~\citep{breyer2022closed,dengler2023iros,jauhri2024active}, harvesting and prediction of intervention actions in agriculture~\citep{pan2023panoptic,lenz2024iros,yao2024safe}, as well as solving jigsaw puzzles of fragmented frescoes in archaeology~\citep{tsesmelis2024NeurIPS}.
For these applications, high-fidelity 3D object representations are critical to enable precise action execution and informed decision-making.
When deployed in initially unknown environments, robots are often required to autonomously reconstruct 3D models of objects to understand their geometries, textures, positions, and orientations before taking action. 

Generating these models typically involves capturing data from multiple viewpoints using onboard sensors such as RGB or depth cameras.
Data acquisition solely following predefined or randomly chosen sensor viewpoints is inefficient, as these approaches fail to adapt to the geometry and spatial distribution of the object to be reconstructed. This can lead to inferior reconstruction results, especially when objects are complex and contain self-occlusions.
To address this, we propose using active reconstruction strategies, where object-specific sensor viewpoints are planned for data acquisition to achieve high-quality 3D object reconstruction.
The key aspect of active reconstruction is view planning for generating viewpoints~\citep{zeng2020cvm} that enables the robot to acquire the most informative sensor measurements.

Existing view planning methods often utilize an iterative paradigm, where the robot incrementally selects the next-best-view~(NBV) based on the current state of the reconstruction~\citep{isler2016icra, palazzolo2018drones, zeng2020iros, mendoza2020prl, pan2022eccv, sunderhauf2023icra, menon2023iros, Jiang2023FisherRF, stachniss2005information, makarenko2002experiment}. 
While these methods progressively refine the object model, they are not without limitations. 
The main drawback is that they generate only a local path to the next viewpoint, without considering globally distributing movement resources, such as the robot’s travel budget or energy constraints. 
Additionally, these approaches often rely on frequent map updates to plan subsequent views, which can be computationally expensive and inefficient.
This iterative process poses challenges for robotic applications, where efficiency and careful resource management are important.

Recent advances in one-shot view planning strategies address the limitations of iterative methods~\citep{pan2022ral1, pan2024icra, hu2024icra, pan2024tro}.
Given initial measurements of an object to be reconstructed, one-shot view planning methods predict an optimized set of viewpoints in a single planning step.
This enables computing the globally shortest path that connects these viewpoints before data collection.
The robot's sensor then follows the pre-planned path to collect measurements without online replanning.
After the data collection is complete, the accumulated measurements are used for object reconstruction in an offline manner.
By decoupling the processes of data collection and object reconstruction, one-shot view planning eliminates the need for iterative updates of the 3D mapping during the mission.
This can allow the robot to use its time and movement resources more efficiently, maximizing online operational efficiency.

To enable one-shot view planning in an initially unknown environment, prior knowledge about the object must be provided to initialize the planning process.
Previous works use planning priors derived from deep neural networks trained on datasets specifically designed for depth sensors~\citep{pan2022ral1, hu2024icra, pan2024tro} or datasets specifying the required number of views for RGB-based reconstruction~\citep{pan2024icra}.
While effective in certain scenarios, these methods have notable limitations.
Approaches relying on depth cameras are sensitive to environmental factors such as lighting conditions, reflective surfaces, and sensor noise, showing limitations in practical applications.
Conversely, leveraging RGB-based 3D reconstruction like neural radiance fields~(NeRFs)~\citep{mildenhall2020eccv} has shown the ability to address such issues effectively.
The RGB-based one-shot view planning method PRV~\citep{pan2024icra} predicts the required number of views for reconstructing objects using NeRFs.
However, this approach can only outputs uniform distributions of viewpoints around the object, without considering its heterogeneous geometry and texture.
As a result, it cannot plan object-specific viewpoints to investigate areas with complex features or occlusions.

Integrating modern 3D diffusion models into one-shot view planning frameworks for RGB-based reconstruction has introduced new possibilities to address these limitations.
Recently, 3D diffusion models have emerged as a powerful tool for generating 3D content from text prompts or images, showcasing their versatility and potential for diverse applications.
These models~\citep{long2022eccv, long2023arxiv, liu2023neurips} are trained on large-scale datasets to learn extensive prior knowledge about the geometry and appearance of objects commonly encountered in real life.
However, recovering an accurate 3D representation from RGB images remains an inherently ill-posed problem, as it allows for multiple plausible interpretations of the object's structure.
Consequently, the 3D models generated by these diffusion techniques often approximate the object's information rather than capturing its true shapes and textures.
This makes them unsuitable for direct application in robotic tasks that demand exact and reliable 3D representations for precise downstream operations, such as manipulation or inspection.
Despite these limitations, our key insight is to leverage the generative power of 3D diffusion models as useful prior information for guiding view planning.

Our prior conference paper introduced DM-OSVP~\citep{pan2024iros}, the first work to leverage 3D diffusion models priors for one-shot view planning in RGB-based reconstruction.
In this previous work, we used 3D diffusion model to generate a mesh of the object from a single RGB image, serving as a proxy for the inaccessible ground truth 3D model and forming the prior for one-shot view planning. Given the proxy model, we proposed formulating one-shot view planning as a customized set covering optimization problem, aiming to determine the minimum set of viewpoints that densely cover the geometries of the generated 3D mesh.
The robot then follows a globally shortest path to efficiently collect informative RGB images around the object.
Once the data collection is complete, the acquired RGB images are used to train a NeRF, enabling the reconstruction of a high-quality object 3D representation.

Our proposed approach in this work follows the idea of using priors from 3D diffusion models to enable one-shot view planning with an object-specific view configuration for RGB-based object reconstruction, illustrated in Fig.~\ref{fig_paper_cover}. 
Building on the limitations identified in our previous method~\citep{pan2024iros}, our novel approach offers significant improvements in terms of applicability of our system in real-world scenarios.
The improvements of this work can be summarized as follows:

\begin{itemize}[leftmargin=*, noitemsep, topsep=0pt]
\item In contrast to DM-OSVP, which relies solely on geometric priors of the generated 3D mesh, our approach additionally incorporates texture complexity analysis in our optimization, resulting in more informative view planning.
\item In comparison to DM-OSVP, which uses a single-image-to-3D diffusion model, our approach employs a multi-image-to-3D diffusion model to provide a more accurate proxy model, addressing the failure cases caused by hallucinations under single-image inputs as mentioned in DM-OSVP.
\item In comparison to DM-OSVP, which assumes a fixed object location and a known object size, we employ the Voxel Carving algorithm~\citep{laurentini1994tpami} to perform a rough localization and estimation of the object. This step enables the dynamic placement of the object-centric view space, allowing our approach to adapt to varying object sizes and placements in the robot workspace.
\item A more comprehensive evaluation of our system, demonstrating a broader application scenario: (a) using different RGB-based reconstruction approaches, including Instant-NGP~\citep{muller2022tog}, Neural Surface (NeuS2)~\citep{wang2023neus2} and 2D Gaussian Splatting (2DGS)~\citep{huang20242d}; (b) assessing both image and mesh metrics; and (c) testing constrained view planning due to unreachable viewpoints in robot's workspace.
\end{itemize}

We conduct extensive experiments on a publicly available object dataset as well as in real-world scenarios to comprehensively evaluate the performance of our approach.
These experiments demonstrate the broad applicability and strong generalization capabilities of our flexible one-shot view planning framework.
By adapting to diverse object geometries and environmental conditions, our method effectively places object-specific viewpoints, ensuring that critical regions of the object are adequately observed while maintaining data collection efficiency.
Compared to baseline methods, our approach achieves a better trade-off between movement cost and reconstruction quality.
This improvement is particularly evident in scenarios with complex object shapes and textures or constrained robot movement spaces.
Our demo video showcasing the capabilities of our approach can be accessed at: \url{https://youtu.be/GBBCn28v-lQ}. 
To further support reproducibility and future research, our implementation will be open-sourced at: \url{https://github.com/psc0628/DM-OSVPplusplus}.

\section{Related Work} \label{sec:RelatedWork}

In this section, we introduce relevant works on view planning for object reconstruction and 3D diffusion models.

\subsection{View Planning for Object Reconstruction} 
\label{sec:VPforOR}

View planning for 3D object reconstruction has been extensively studied and remains a promising area of research in robotics~\citep{tarabanis1995survey,scott2003view, stachniss2005information,chen2011active,zeng2020cvm,maboudi2022review,popovic2024learning}.
As categorized by~\citep{pan2024tro}, view planning methods can be broadly divided into next-best-view planning and one-shot view planning approaches.

\subsubsection{Next-Best-View Planning}
 
Without any prior knowledge, a common approach is to iteratively plan the NBV based on the current reconstruction state, thereby maximizing the information about the target object in a greedy manner.
One branch of NBV planning requires a depth sensor to collect the information of the object.
\citet{isler2016icra} propose an NBV selection strategy that evaluates information gain by considering both visibility and the likelihood of observing previously unobserved regions of the target object in voxel-based representation.
Similarly, \citet{pan2022ral2, pan2023cviu} assign weights to the 3D space based on the visibility and proximity to observable surfaces, followed by coverage optimization to determine the viewpoint.
\citet{menon2023iros} extend information gain by introducing a shape completion framework that predicts missing object surfaces based on partial observations and employs NBV planning to maximize coverage of the estimated geometry.
PC-NBV~\citep{zeng2020iros} employs a neural network trained to predict the utility of candidate views using partial point clouds.

Another line of research in NBV planning focuses on RGB-based 3D object reconstruction.
\citet{lin2022rssworkshop} solves the NBV planning by training ensembles of NeRF models, utilizing the ensemble's variance to measure model uncertainty.
\citet{sunderhauf2023icra} enhances the ensemble-based uncertainty by integrating the density information.
\citet{jin2023iros} integrate uncertainty estimation into image-based neural rendering to guide NBV selection in a mapless way. 
\citet{Jiang2023FisherRF} leverage Fisher Information to directly quantify the observed information on the parameters of radiance fields.
\citet{jin2025activegs} propose the confidence modeling for the Gaussian splatting to identify under-reconstructed areas.

\subsubsection{One-Shot View Planning}

Although NBV planning demonstrates promising results in active object reconstruction, it often relies on computationally intensive online map updates, and its greedy nature results in inefficient exploration paths.
To overcome these limitations, recent works propose the one-shot view planning paradigm~\citep{pan2022ral1,hu2024icra,pan2024tro,pan2024icra,pan2024iros,isaacjose2025govmp}. 
Given initial measurements, one-shot view planning predicts all necessary views at once and computes the globally shortest path connecting them, thereby minimizing the movement costs.
The previous work SCVP~\citep{pan2022ral1} trains a neural network in a supervised manner and directly predicts an effective global view configuration based on the initial point cloud.
To generate training labels, the authors solve a set covering problem to determine a view configuration that fully encompasses the ground truth 3D models.
Building upon this, \citet{hu2024icra} enhance efficiency by integrating point cloud-based implicit surface reconstruction.
This approach completes missing surfaces prior to executing one-shot view planning, thereby reducing the number of required views for comprehensive 3D reconstruction.

In the field of RGB-based one-shot view planning, \citet{pan2024icra} introduce a view prediction network that estimates the number of views needed to achieve the performance ceiling of NeRF-based reconstruction.
Since this approach relies on a fixed pattern for distributing views, it cannot adjust the configuration to accommodate varying object-specific geometries and textures, such as allocating a higher number of views to more complex parts of the scene.
Our previous work DM-OSVP~\citep{pan2024iros} addresses this issue by exploiting the proxy geometry generated by 3D diffusion models and planning view configurations specifically associated with the objects.
In this work, we further extend the flexibility of the DM-OSVP under diverse
conditions and real-world environments.

\subsection{3D Diffusion Models}
\label{sec:3DDiff}

Diffusion models, which utilize stochastic processes to effectively capture complex data distributions, have emerged as state-of-the-art generative models, excelling in image-related tasks by iterative denoising process. 

The remarkable success of 2D image generative diffusion models has catalyzed a growing interest in extending their capabilities beyond 2D imagery, with increasing attention directed toward the development of 3D object diffusion models. 
The earliest studies focused on text-to-3D generation, directly trained on 3D paired data~\citep{chen2019accv, mittal2022cvpr, fu2022nips, vahdat2022neurips} or unlabeled 3D data~\citep{sanghi2022cvpr, sanghi2022arxiv}.
Another line of text-to-3D generation methods leverage large pretrained 2D text-to-image diffusion models for per-shape optimization~\citep{poole2023iclr, wang2023neurips} or in a feed-forward manner~\citep{li2023arxiv}.

\subsubsection{Single-Image-to-3D}
While text-guided 3D object generation has shown promising results, the inherent limitations of text descriptions, such as their inability to capture detailed visual attributes or precise spatial information, have motivated recent advancements in image-based 3D generation.
One branch of existing work explores 3D object generation from a single image.
Zero-1-to-3~\citep{liu2023iccv} pioneers open-world single-image-to-3D generation by utilizing zero-shot novel view synthesis.
As Zero-1-to-3 generates images in different poses separately, results for the same object are inconsistent in geometry and texture.
Recent approaches such as One-2-3-45~\citep{liu2023neurips}, SyncDreamer ~\citep{liu2024iclr} and Consistent123~\citep{lin2024acmmm} extend Zero-1-to-3 with additional layers to acquire more 3D-consistent results.
However, their performance is yet limited by the inconsistency between multi-view images.
The follow-up work One-2-3-45++~\citep{liu2024cvpr} mitigates the problem of inconsistencies by conditioning the multi-view image generation on each other.
This method exploits the generated multi-view consistent images as the guidance for 3D diffusion to directly produce high-quality meshes in a short time.
However, generating 3D objects from a single image is still prone to hallucinations and wrong shape prediction.

\subsubsection{Multi-Image-to-3D Generation}
Recent advances in image-to-3D generation methods address the limitations of hallucinations inherent in single-image inputs by conditioning on multi-image inputs.
The state-of-the-art multi-view conditioned diffusion model EscherNet~\citep{kong2024cvpr} integrates a specialized camera positional encoding, allowing it to generate an arbitrary number of target views at any camera pose, based on any number of reference views on an object-centric sphere.
While this integration effectively reduces hallucination issues, it remains unsuitable for precise and reliable robotic operations due to residual inconsistencies in 3D generation.
These inconsistencies arise because the model generates 2D images as an intermediate step, which can compromise the overall 3D coherence.
Instead, we use it as promising priors for our one-shot view planning.

\section{One-Shot View Planning as Set Covering Optimization} 
\label{sec:OSVPlanning}

The foundation of our active RGB-based 3D~object reconstruction approach is one-shot view planning, which plans a global path consisting of multiple viewpoints at once.

Given a discretized view space $\mathcal{V}\subset\mathbb{R}^3\times SO(3)$, which represents a dense set of candidate viewpoints, the goal of one-shot view planning is to identify the minimum subset of selected views $\mathcal{V}^\star\subset\mathcal{V}$ that densely cover the 3D mesh of the object of interest generated by a 3D diffusion model. 
Such a one-shot view planning process can be formulated as a set covering optimization problem~\citep{pan2022ral1, pan2024icra, hu2024icra, pan2024tro}.
We detail how to adapt the conventional set covering optimization problem specifically for RGB-based reconstruction methods, ensuring a dense object coverage for high-quality reconstruction.

To facilitate the efficiency of set covering optimization, sparse surface representations are desired. 
To this end, we first sample a set of surface points from the mesh produced by the 3D diffusion model and subsequently voxelize them using OctoMap~\citep{hornung2013ar} to get a sparse surface point set $\mathcal{P}_{\mathit{surf}}$, with surface point $p_i \in \mathcal{P}_{\mathit{surf}}$.
We denote $v$ as a candidate view within the aforementioned discrete view space $\mathcal{V}$ and $\mathcal{P}_v$ as the set of surface points observable from this view.
Each set $\mathcal{P}_v$ is determined via the ray-casting process implemented in OctoMap.
We define an indicator function $I(p,v)$ to represent whether a surface point $p$ is observable from the candidate view $v$:

\begin{equation}
\label{equ:indicator}
    I(p,v) = 
    \begin{cases}
        1   & \text{if } p \in \mathcal{P}_v \\
        0   & \text{otherwise}
    \end{cases}
    \, .
\end{equation}

Given $\mathcal{P}_{\mathit{surf}}$ and each $\mathcal{P}_v$, the conventional set covering optimization problem aims to find the minimum set of views required for completely covering the surface points.
For instance, consider $\mathcal{P}_{\mathit{surf}} = \left\{p_1,p_2,p_3\right\}$, $\mathcal{P}_{v_1} = \left\{p_1,p_2\right\}$, $P_{v_2} = \left\{p_2,p_3\right\}$, and $\mathcal{P}_{v_3} = \left\{p_1,p_3\right\}$. The union of these three sets equals the entire surface set, i.e., $\bigcup_{v}\mathcal{P}_v = \mathcal{P}_{\mathit{surf}}$. However, we can cover all surface points with only two sets, $\mathcal{P}_{v_1}$ and $\mathcal{P}_{v_2}$.

\subsection{Multi-View Constraints}

\begin{figure}[!t]
\centering
\includegraphics[width=1.0\columnwidth]{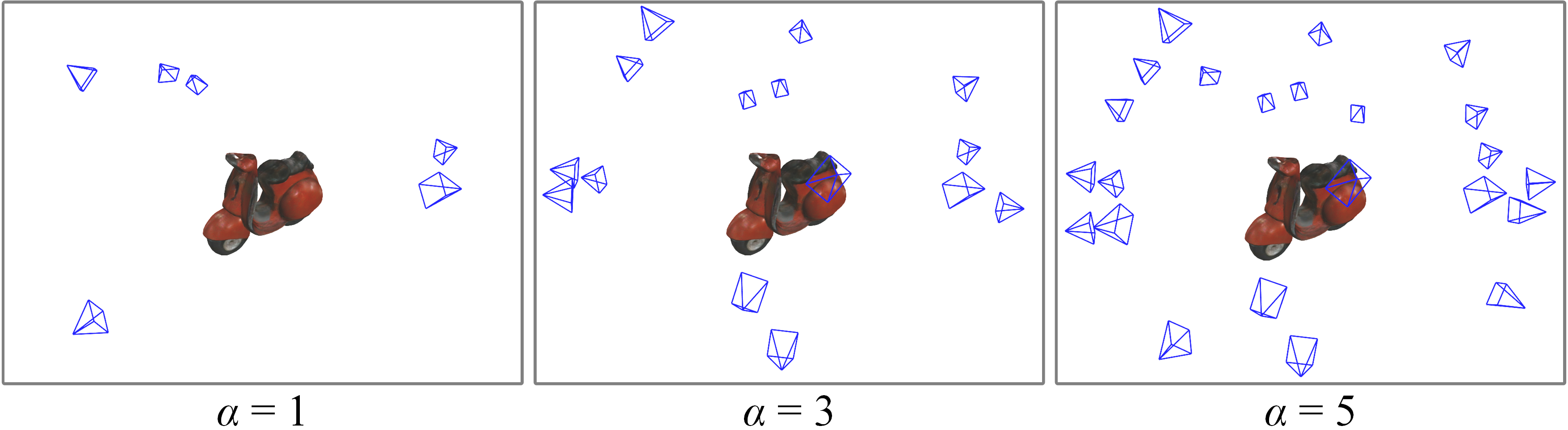}
\caption{
Illustration of the impact of multi-view constraints. $\alpha$ denotes the minimum number of views required to observe each surface point. Larger $\alpha$ values lead to optimization solutions with more views densely covering the surfaces.} 
\label{fig_miniuxm_covering}
\end{figure}

The set covering optimization problem introduced above requires that each surface point should be covered by at least one view from the discrete view space $\mathcal{V}$.
This definition aligns well with object reconstruction employing depth-sensing modalities~\citep{pan2022ral1, pan2024tro, hu2024icra}, as surfaces can be recovered by direct depth fusion when provided with a corresponding point cloud observation.
However, for RGB-based object reconstruction, map representation learning can be achieved by minimizing the photometric loss when reprojecting hypothetical surface points back to 2D image planes, which requires that a surface point should be observed from different perspectives to recover its true 3D representation.
This implies that planned views covering all surface points of the generated mesh once are not sufficient for RGB-based object reconstruction.

Therefore, we customize the set covering optimization problem for RGB-based object reconstruction.
Rather than requiring each surface point to be observed by at least one view, we propose multi-view constraints to enforce that a given surface point should be covered by a minimum number $\alpha \in \mathbb{N}^{+}$ of views to account for multi-view learning.
Larger~$\alpha$ values require denser surface coverage in our optimization problem, resulting in solutions with more views required as shown in Fig.~\ref{fig_miniuxm_covering}.
Note that when $\alpha \geq 2$, we exclude points that are visible from fewer than $\alpha$ views.
This mechanism ensures the optimization problem has a feasible solution.

\subsection{Object Complexity Analysis}

However, considering surface point coverage with multi-view constraints $\alpha$ accounts only for geometric information, neglecting the texture information of the point.
Unlike our previous version~\citep{pan2024iros}, which relies solely on the raw geometric covering, one key insight of our novel approach is to integrate complexity analysis of each surface point $p$ into our multi-view constraints to enable a more informative and comprehensive coverage. 

\begin{figure}[!t]
\centering
\includegraphics[width=0.95\columnwidth]{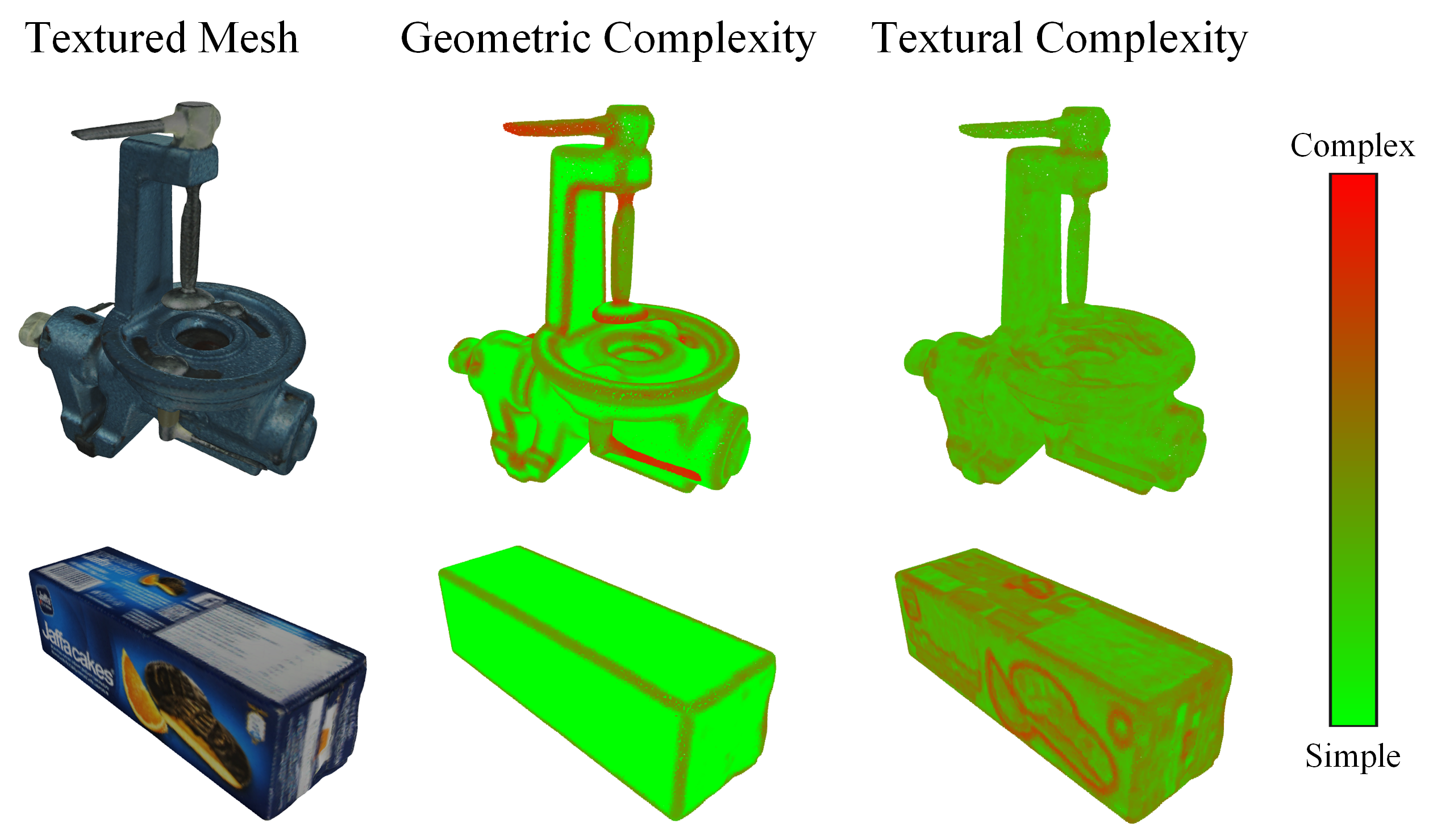}
\caption{
Illustration of the geometric and textural complexity of two representative objects: one with geometric complexity and simple texture (top), and the other with simple geometry and complex texture (bottom).
Regions with greater complexity are characterized by dramatic changes, such as significant curvature variations or pronounced color transitions.
} 
\label{fig_complexity}
\end{figure}
 
We define the point-level complexity based on both local geometric and textural information of a point to its neighboring points. Specifically, we utilize the PFHRGB feature, an extension of Point Feature Histograms (PFH)~\citep{rusu2008aligning}, which also incorporates color information and has demonstrated leading performance in benchmarks~\citep{alexandre20123d}.
The PFHRGB feature considers surface normals and curvature estimates for capturing geometric information, as well as color values for representing texture information.
It computes angular variations derived from the normals between the surface point $p$ and its neighboring points, storing these in a 125-bin histogram to represent geometric information.
Additionally, it calculates color ratios across all three RGB channels between the surface point $p$ and its neighboring points, constructing a separate 125-bin histogram to capture textural information.

The histograms are well-designed to effectively quantify the complexity of a surface point.
In a simple case, such as when normals and colors are consistent around a point, the histogram values will be highly concentrated around one or a few bins.
Conversely, in a complex case, such as when there are significant variations in curvature and color, the histogram values will be more evenly distributed across all bins.
Therefore, we propose defining the complexity $C(p)$ of a surface point $p$ using entropy as a measure:

\begin{equation}
\label{equ:complexity}
C(p) = -\sum^{125}_{i}{q_i\ln\left(q_i\right)} -\sum^{125}_{j}{q_j\ln\left(q_j\right)},
\end{equation}
where $i$ represents each geometric bin and $j$ represents each textural bin, while $q_i$ and $q_j$ denote the probabilities calculated as the count of the respective bin divided by the total count of all bins in the histogram.
We normalize $C(p)$ to the range $(0,1]$ to ensure a unified complexity among different objects. Fig.~\ref{fig_complexity} illustrates the complexity analysis of objects with varying characteristics.

To integrate $C(p)$ into our multi-view constraints, we require each surface point to be covered at least $\left\lceil \alpha C(p) \right\rceil$ times, where $\alpha$ determines the coverage requirements for the most complex points, which must be observed at least $\alpha$ times.
This approach allows the multi-view constraints to adapt to the local complexity of the object, providing constraints of varying intensities based on the specific geometric and textural intricacies.

\subsection{Distance Constraints}

\begin{figure}[!t]
\centering
\includegraphics[width=0.9\columnwidth]{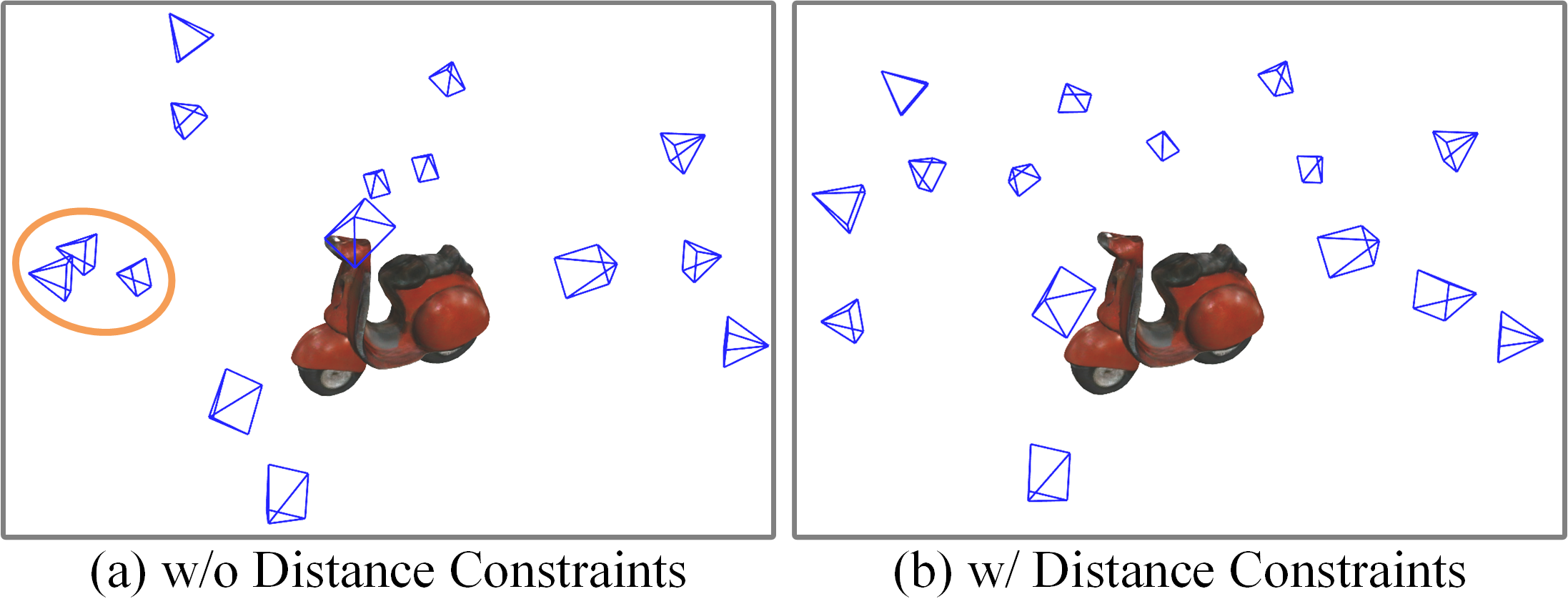}
\caption{
Illustration of the impact of distance constraints: (a) spatially clustered views (the orange circle showcases an example of clustered views); (b) spatially more uniform views. Both view configurations are feasible solutions. By incorporating distance constraints, we express the preference for spatially uniform distribution to avoid redundant information in clustered view configurations.} 
\label{fig_spatial_constraints}
\end{figure}

Moreover, our multi-view covering setup may contain multiple feasible solutions since most of the surface points can be observed from a large range of view perspectives.
Some of them lead to views clustered closely together in Euclidean space.
Fig.~\ref{fig_spatial_constraints}(a) illustrates an instance of spatially clustered views for an example object.
These spatially clustered views exhibit similarity in the collected images, thus leading to redundant information about the object.

To alleviate this issue, we introduce a parameter $\beta \in \mathbb{R}^{\geq 0}$ for additional distance constraints to avoid selecting spatially clustered views.
We denote $d_v^{v'}$ as the Euclidean distance between views $v$ and $v'$, while $d_v^{min}$ is the Euclidean distance from view $v$ to its nearest neighboring view.
We prevent other views within a specific distance~$\beta \, d_v^{min}$ of the view~$v$ from being selected again in the solution.
A larger $\beta$ leads to more spatially uniform views, while an excessively large value can render the problem infeasible.
For our view planning, we try to find the maximum $\beta$ value that still yields an optimization solution. 
For instance, if $\beta$ approaches infinity, selecting one view would prevent the selection of any additional views. 
Given that different objects exhibit diverse geometries, their respective maximum $\beta$ values also vary.
Therefore, we run optimization iteratively to find the maximum $\beta$ for a specific object in an automatic manner.
Specifically, we adopt binary search in our implementation to find out the object-specific $\beta$ that still yields a feasible optimization solution with a search step of 0.1.
We present an instance solution in Fig.~\ref{fig_spatial_constraints}(b) showcasing the optimized minimum set of views required for densely covering the object with distance constraints.

\subsection{Integer Linear Programming Formulation}

Taking all these conditions into account, we formulate our set covering optimization problem as a constrained integer linear programming problem defined as follows:

\vspace{-0.2cm}
\begin{equation}
\label{equ:ILP}
\begin{aligned}
\min: \, & \sum_{v\in \mathcal{V}} x_v \, ,\\
\mathrm{s.t.}: \,
& (a) \, x_{v} \in \{0,1\}   && \forall v \in \mathcal{V} \\
& (b) \, \sum_{v \in \mathcal{V}} I(p, v)\, x_{v} \geq \left\lceil \alpha C(p) \right\rceil  && \forall p \in \mathcal{P}_{\mathit{surf}} \\
& (c) \, x_v + x_{v'} \leq 1  && \forall d_v^{v'} \leq\beta\, d_v^{min},
\end{aligned}
\end{equation}
where the objective function $\sum_{v \in \mathcal{V}} x_v$ is designed to minimize the total number of selected views, while subject to three constraints: (a) $x_{v}$ is a binary variable representing whether a view $v$ is included in the set of selected views or not; (b) each surface point $p \in \mathcal{P}_{\mathit{surf}}$ must be observed by a minimum of $\left\lceil \alpha C(p) \right\rceil$ selected views; and (c) if a view $v$ is selected, any neighboring view $v'$, whose distance $d_v^{v'}$ is smaller than $\beta \, d_v^{min}$, must not be selected. 
We employ the Gurobi optimizer~\citep{gurobi2021gurobi}, a linear programming solver to compute the solution for the problem. 
After solving the problem, we find all $x_v=1$ to obtain the minimum subset $\mathcal{V}^\star$ of selected views.

\section{View Generation and Object Reconstruction}
\label{sec:SO}

\begin{figure*}[!t]
\centering
\includegraphics[width=1.0\textwidth]{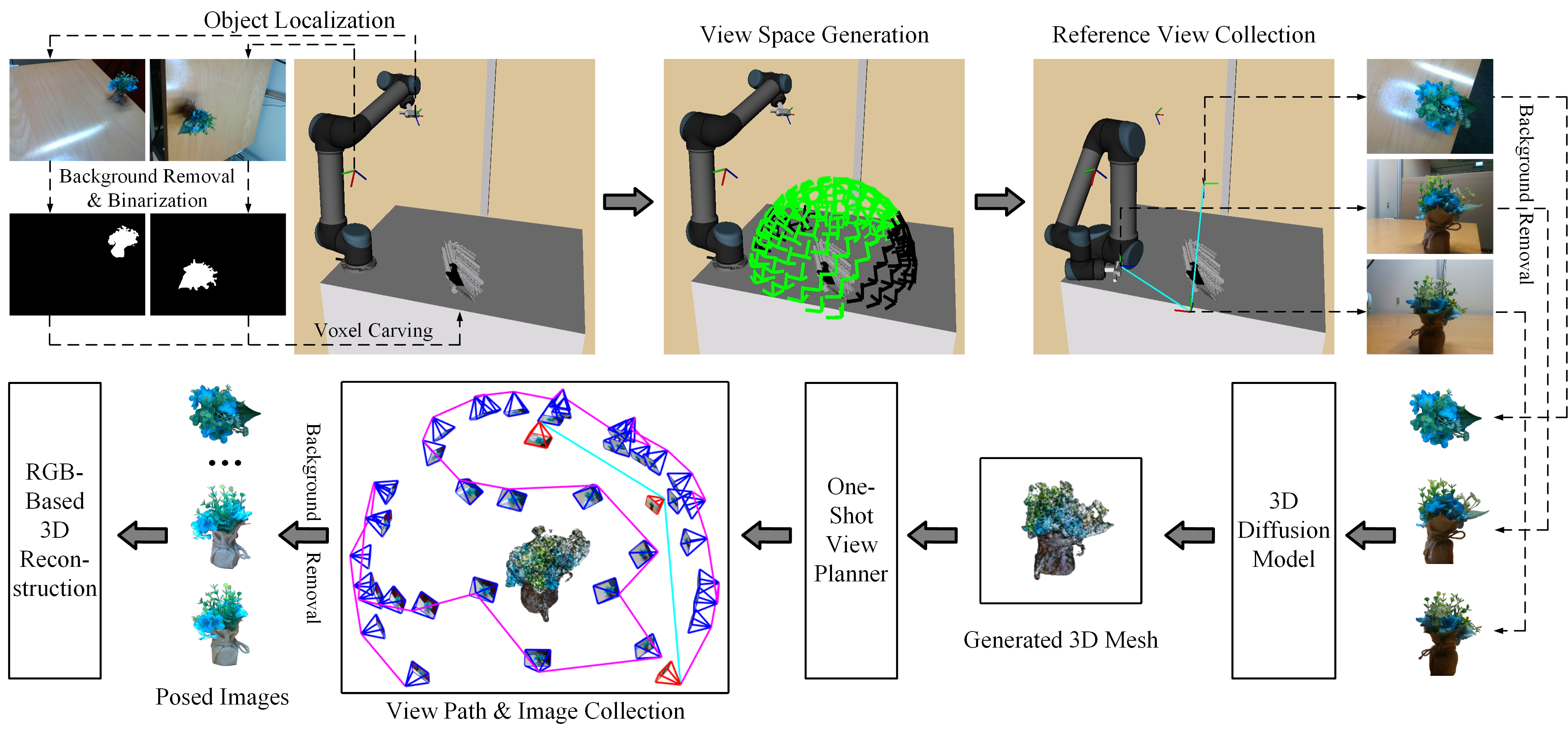}
\caption{An overview of our proposed active RGB-based 3D reconstruction system utilizing one-shot view planning with 3D~diffusion models.
Starting with two RGB images capturing an overview of the tabletop, the object to be reconstructed is roughly localized using the Voxel Carving method~\citep{laurentini1994tpami}.
Given the localized carved mesh, we generate an object-centric view space and evaluate the reachability of each viewpoint with MoveIt~\citep{chitta2012moveit}, where green indicates reachable viewpoints, while black points represent non-reachable ones.
Three reference view images (top, left, front) are collected as the input of the 3D diffusion model EscherNet~\citep{kong2024cvpr} to generate a 3D mesh, which serves as a proxy to the ground truth 3D model and is the basis for our one-shot view planner.
Based on this prior, we construct the one-shot view planning task as a customized set covering optimization problem and solve it to obtain a minimum set of views required to densely cover the mesh surfaces.
The RGB camera follows the generated globally shortest path (purple) to collect posed RGB images, which we use to obtain the 3D representation of the object after the data acquisition is completed.
}
\label{fig_pipeline}
\end{figure*}

In this work, we consider reconstructing a high-quality 3D representation of an object in an unknown and spatially limited 3D space through a sequence of planned camera views navigated by a robotic arm.
The robotic arm is equipped with a calibrated RGB camera at its end effector to acquire posed images.
A view in the candidate view space of the eye-in-hand camera is defined by its 3D position and view direction.
The view planning problem is to generate the optimized sequence of views that simultaneously satisfies the following two objectives: reconstruction of a high-quality 3D representation, and efficient collection of these images within short paths.
To solve this problem, we use the aforementioned novel one-shot view planning method with a 3D diffusion model to provide a proxy 3D mesh for optimization.
An overview of our system is shown in Fig.~\ref{fig_pipeline}.

\subsection{Object Localization}

In an initially unknown environment, we first need to localize the object and generate view space around it.
To localize the object on the tabletop, we utilize the Voxel Carving algorithm\footnote{\url{https://github.com/unclearness/vacancy}}, which provides a rough estimation of the object to be reconstructed~\citep{laurentini1994tpami}.

In particular, we define two fixed viewpoints within the robot workspace to capture an overview of the tabletop.
The background of the collected images is removed using the Rembg tool\footnote{\url{https://github.com/danielgatis/rembg}} with the pertrained BiRefNet-general~\citep{zheng2024CAAI}.
The processed images are then binarized to obtain silhouettes, which are used to back-project generalized view cones that enclose the object.
The voxelized bounding box of the tabletop space is then carved by intersecting these projected zones.
The algorithm finally generates a carved mesh using the marching cubes method, serving as a bounding geometry for the object.
The two fixed viewpoints are referred to as carving views and the number of carving views required is analyzed in Sec.~\ref{subsebsec:CarvingViews}.

\subsection{View Space Generation}

Once the carved mesh is obtained from the localization module, we calculate the object's rough center and size.
The center is defined as the centroid of the surface points on the carved mesh, while the size is determined as the radius of the minimum bounding sphere enclosing the carved mesh.
We then position the center of the view space at the object's center, with a radius set to three times the object size.
This configuration ensures that the object remains within the camera's field of view, adjustable based on different cameras.

We sample a set $\mathcal{V}\subset\mathbb{R}^3\times SO(3)$ of dense hemispherical views evenly distributed around the placed object by solving the Tammes problem~\citep{lai2023iterated}.
This discretization of object-centric view space is suitable for object reconstruction tasks, with view direction pointing to the center of the object, as validated in~\citep{pan2024tro,kong2024cvpr}.

Moreover, as the robot's workspace is constrained by physical limitations, viewpoints out of robot's workspace or in occupied space are unreachable.
To address this, we compute the reachability of each sampled candidate viewpoint within the discretized view space using MoveIt~\citep{chitta2012moveit}.
Only the reachable viewpoints are retained for the subsequent view planning process.

\subsection{Priors from 3D Diffusion Model}

Another key module of our system is a 3D diffusion model that predicts the corresponding mesh from a few reference RGB images as initial observations, serving as a basis for our set covering optimization.
Specifically, we adopt the multi-image-to-3D diffusion model EscherNet~\citep{kong2024cvpr}, replacing the One-2-3-45++~\citep{liu2024cvpr} used in our previous version.
The switch was made because One-2-3-45++ supports only single-image input, leading to hallucinations under such conditions, as noted in DM-OSVP~\citep{pan2024iros}.
Additionally, it lacks crucial scale and position information about the object and input viewpoints, limiting its real-world applications.
In contrast, EscherNet encodes viewpoint information directly into its inputs along with images, enabling more practical and reliable performance for 3D model generation.

The reference views provided to EscherNet are captured from three key perspectives: top, left or right, and front or back along the object's local axes.
In case of unreachability, we select the closest candidate view to the predefined ones in the view space as the reference views.
These views capture essential information about the object's geometry and texture on the tabletop.
The impact of the number of reference views is analyzed in the Sec.~\ref{subsec:ReferenceViews}.

Since EscherNet outputs a set of 2D images of the object, we follow the 3D generation process outlined in their paper~\citep{kong2024cvpr}.
Specifically, 36 fixed views are generated to train a NeuS representation for obtaining the mesh.
To enable faster inference, we utilize NeuS2~\citep{wang2023neus2} instead.
EscherNet is trained on Objaverse~\citep{deitke2023cvpr}, a large-scale 3D model dataset, to learn prior knowledge of commonly seen objects, demonstrating strong generalization capabilities on unseen objects.
Leveraging this powerful tool, the generated meshes are used as priors for our one-shot view planning.

\subsection{View Path Generation}

By planning all required views before data collection, the one-shot view planning paradigm shows a major advantage in reducing movement costs.
Given the optimized set $\mathcal{V}^\star$ of views from our customized set covering optimization, we plan the globally shortest path connecting all views by solving the shortest Hamiltonian path problem on a graph, which is similar to the traveling salesman problem~\citep{osswald2016ral}.
During online data collection, the robot's RGB camera follows the global path to acquire posed RGB measurements at planned views.
We follow the point-to-point local path planning method~\citep{pan2024tro} to avoid collisions with the object.

\subsection{Object Reconstruction}

After data collection, we utilize an RGB-based reconstruction method to obtain the final 3D representation of the object.
Specifically, we adopt Instant-NGP~\citep{muller2022tog} for training a NeRF representation using these posed RGB images.
It integrates joint pose optimization during the training process.
This feature is particularly valuable in real-world scenarios, where the noise of recorded view pose is unavoidable due to calibration errors.

\section{Experimental Results} \label{sec:Experiments}

Our experiments are designed to demonstrate that: (1) object complexity analysis enables a more informative set covering optimization; (2) the multi-image-to-3D diffusion model provides a more accurate 3D mesh for one-shot view planning; (3) leveraging powerful priors from 3D diffusion models achieves a better balance between reconstruction quality and on-board resources compared to baselines across various RGB-based reconstruction methods; and (4) in real-world scenarios, our object localization module and reachability checking effectively provide the necessary information to enable dynamic placement of the object-centric view space for constrained view planning. Additionally, we analyze our multi-view constraints to offer practical guidance for system deployment.

\subsection{Setup and Evaluation}

\begin{figure}[!t]
\centering
\includegraphics[width=0.95\columnwidth]{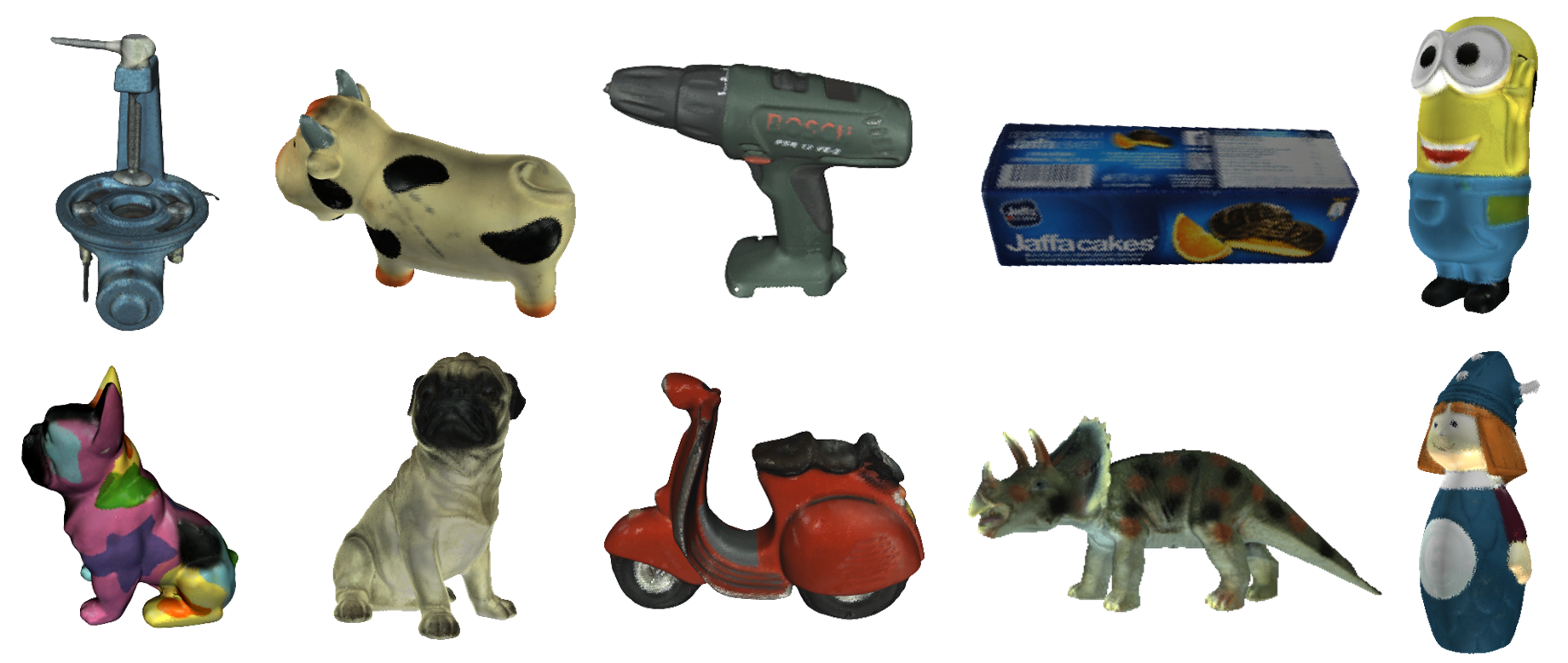}
\caption{
The ten test objects used in our simulation experiments. They have varying geometric and textural complexities.
} 
\label{fig_object_3D_models}
\vspace{-0.2cm}
\end{figure}

In our simulation experiments, we render object RGB measurements using an OpenGL engine with uniform lighting. All RGB images are captured at a resolution of $640 \times 480$ pixels.
We evaluate our approach on ten geometrically or textually complex 3D object models, as shown in Fig.~\ref{fig_object_3D_models}, sourced from the HomebrewedDB dataset~\citep{kaskman2019cvpr}.
These test objects are normalized to fit within a bounding sphere of $1.0$ radius at the center of $(0,0,0)$.
We then use an object-centric hemispherical view space consisting of 144 uniformly distributed candidate views for view planning, with a radius of $3.0$.
For EscherNet reference views, we use the default resolution of $256 \times 256$ pixels, as specified in their codes~\citep{kong2024cvpr}, with the image and camera parameters adjusted accordingly.
The OctoMap grid used in our optimization is set to $50 \times 50 \times 50$ voxels for voxelizing the mesh surface points.
We train and evaluate the methods on a PC with an Intel Core i7-12700H CPU, 32 GB RAM, and an NVIDIA RTX3060 laptop GPU.

We evaluate reconstruction quality using novel view rendering metrics, including Peak Signal-to-Noise Ratio (PSNR), Structural Similarity Index (SSIM), and Learned Perceptual Image Patch Similarity (LPIPS), as well as geometric metrics, including Hausdorff Distance (HD), Chamfer Distance (CD), and Earth Mover's Distance (EMD), to analyze the quality of both image generation and mesh geometries~\citep{mildenhall2020eccv,nehme2023textured}.
We evaluate reconstruction efficiency through inference time for view planning and the cumulative movement cost for data collection, measured in Euclidean distance~\citep{pan2024iros}.
For evaluation, the Instant-NGP~\citep{muller2022tog} training steps are set to 10,000.
The PSNR, SSIM, and LPIPS are averaged on 100 uniform distributed novel views. 
The points sampled on the mesh surface for distance metrics are 10,000 points for HD and CD, and 1,000 points for EMD to ensure computational efficiency.

\subsection{Study on One-Shot View Planners}

\begin{table}[!t]
\centering
\resizebox{1.0\columnwidth}{!}{%
\begin{tabular}{ccccc}
\toprule
Planners       & \makecell{Multiview\\Constraints} & \makecell{Distance\\constraints} & \makecell{Complexity\\Analysis} \\
\midrule
SetRaw         & $\checkmark$                     & ×                    & ×                  \\
SetRaw+Uni     & $\checkmark$                     & $\checkmark$                    & ×                  \\
SetRaw+Uni+Com & $\checkmark$                     & $\checkmark$                  & $\checkmark$                  \\
\bottomrule
\end{tabular}
}
\vspace{0.2cm}
\caption{
Different one-shot view planning setups.
}
\label{tab_planners}
\end{table}

\begin{figure*}[!t]
\centering
\includegraphics[width=0.9\textwidth]{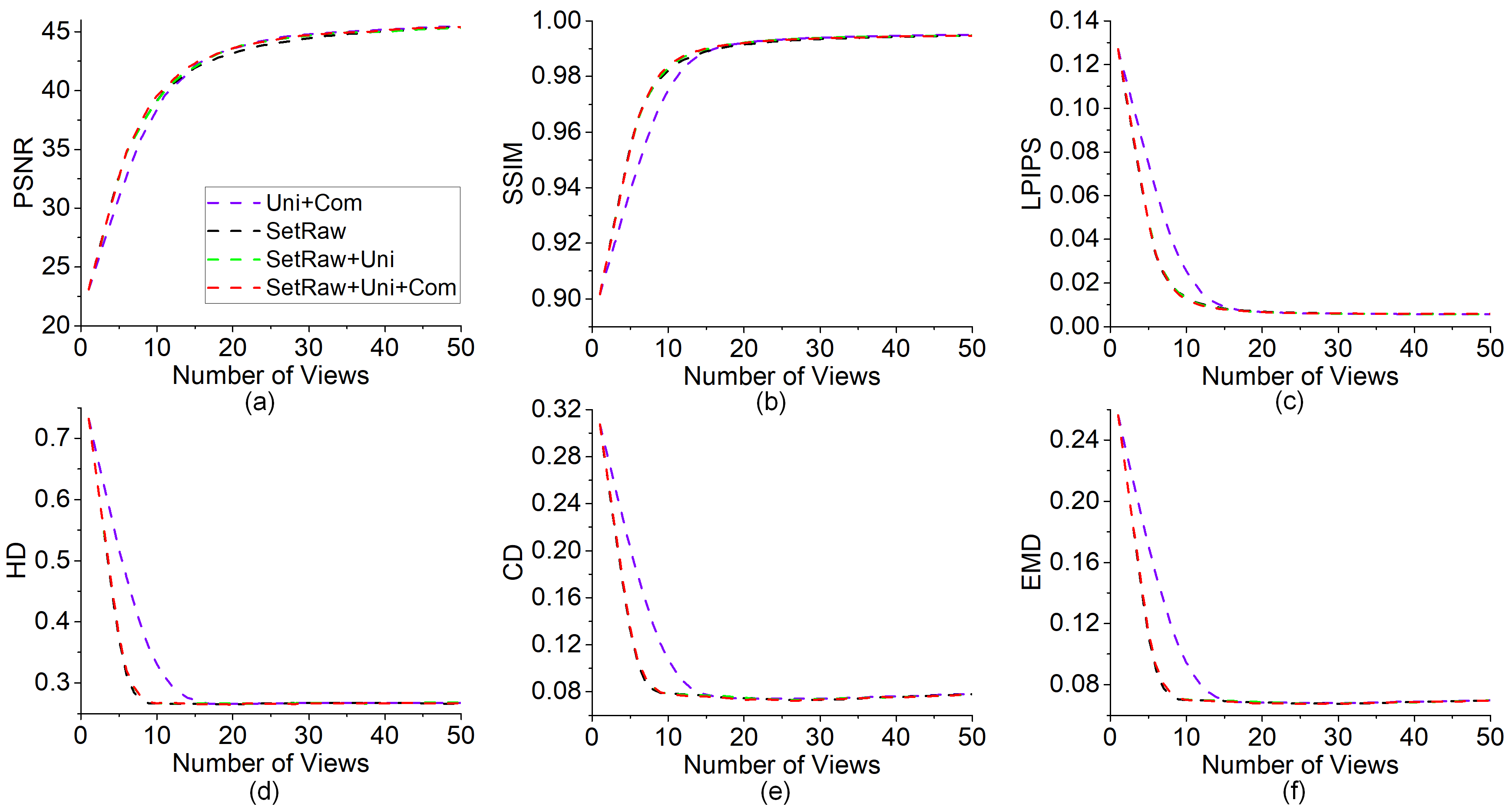}
\caption{
Evaluation of different covering planners on the ground truth meshes of the objects.
The number of views selected by each planner is automatically determined by the multi-view constraints, \ie, $\alpha$ values.
Hence, we run each planner with incrementally increasing $\alpha$ values, starting from 1, until the number of selected views exceeds 50.
Linear interpolation is then applied to draw the curves.
Each curve represents the average performance across 10 test objects.
As can be seen, \textit{SetRaw+Uni+Com} consistently achieves the best performance across all metrics and for all numbers of views. This indicates its ability to identify the most informative views, making it our final choice.
} 
\label{fig_planner_ablation}
\end{figure*}

In this section, we evaluate the effectiveness of customized set covering optimization problems using the ground truth mesh of the objects, ensuring that only the optimization performance is assessed.
We study the ablations on our customized constraints as denoted in Table~\ref{tab_planners}.
Furthermore, we also conduct a comparison to a uniform-aimed covering planner, denoted as Uni+Com.
It selects the most spatial uniform distributed views with complexity analysis, which is detailed in the \autoref{apd:planner_uniform}.

The reconstruction quality of each planner is presented in Fig.~\ref{fig_planner_ablation}.
Among all the planners, SetRaw+Uni+Com achieves the best performance across all metrics, demonstrating that object complexity analysis contributes to a more comprehensive set covering optimization.
Notably, it primarily improves image-based metrics while showing similar performance to SetRaw and SetRaw+Uni in geometric metrics.
This is expected, as texture information does not directly influence geometric reconstruction.
Additionally, the worst performance observed with Uni+Com highlights the importance of object-specific view planning by considering the information contained in the selected viewpoints.
Based on these findings, we select SetRaw+Uni+Com as our planner for the subsequent experiments, as it consistently achieves the best performance across all scenarios, regardless of the number of views selected by the planner.

\subsection{Study on Reference Views to EscherNet}
\label{subsec:ReferenceViews}

\begin{figure*}[!t]
\centering
\includegraphics[width=0.75\textwidth]{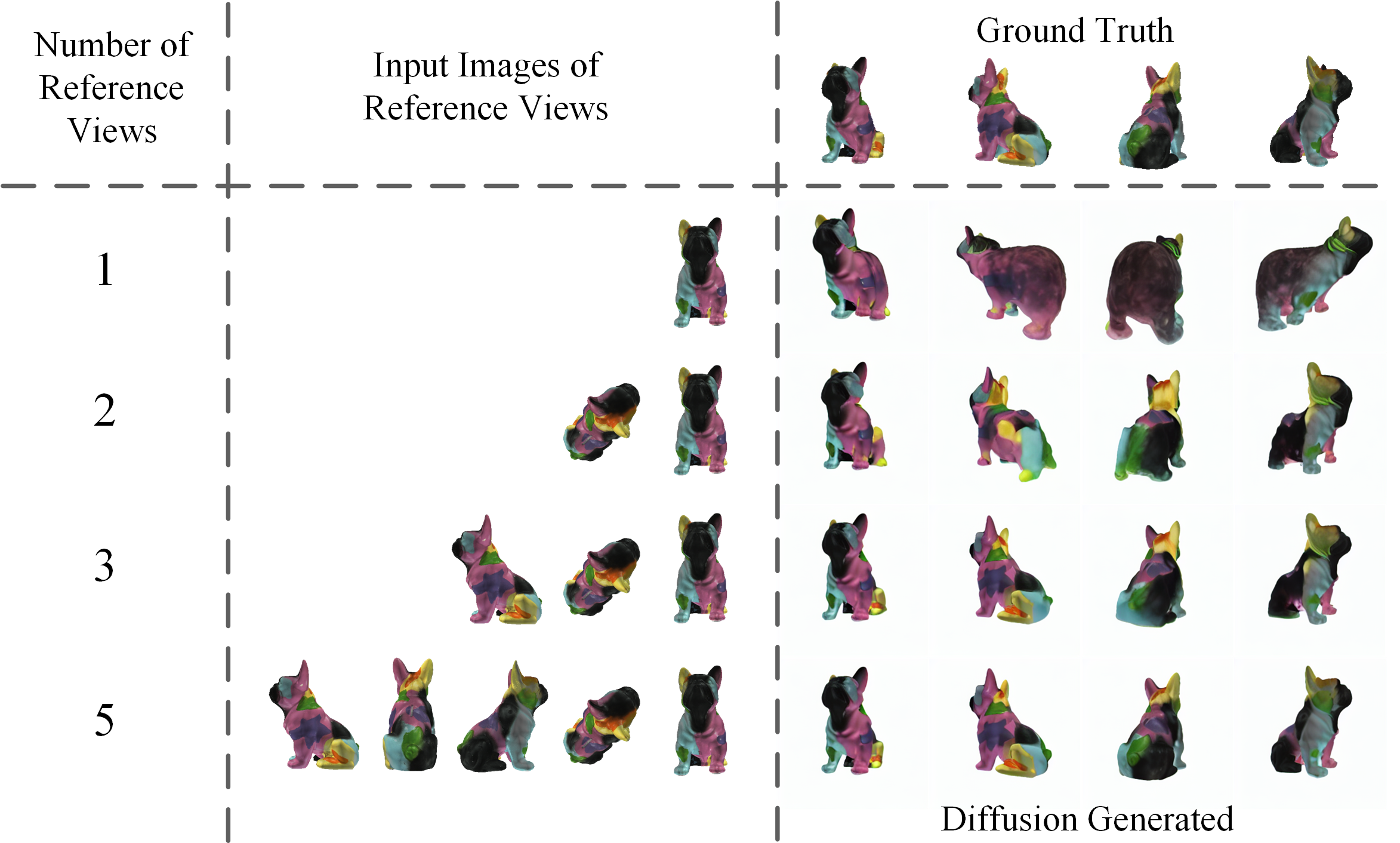}
\caption{
Illustration of the impact of reference views on the quality of the EscherNet-generated 2D images for obtaining the meshes. 
As can be seen, using a single reference view results in significant discrepancies in geometry and texture, and 5 views offer a slight improvement over 3, indicating 3 reference views are sufficient for approximating the ground truth.
} 
\label{fig_reference_views}
\end{figure*}

In this section, we examine the number of reference views needed for EscherNet, the multi-image-to-3D diffusion model, to generate a reliable 3D mesh suitable for our one-shot view planning.
First, we present a qualitative example that visualizes the generated 2D images for obtaining the 3D meshes of the Dog object with different numbers of reference views, as shown in Fig.~\ref{fig_reference_views}.
The results indicate that the quality of the generated meshes improves noticeably when increasing the reference views from 1 to 2, shows some improvement from 2 to 3, but very little improvement when further increasing to 5, aligned with the finding in EscherNet paper.

Second, to further investigate how the quality of the generated meshes impacts our view planner, we provide a quantitative analysis of the differences in the number of views selected by the proposed planner when using ground truth meshes versus EscherNet-generated meshes.
This evaluation is designed because the reconstruction quality metrics are expected to be quite similar when the same number of views is selected by the proposed planner.
Therefore, we focus on how well it approximates the complexity of the ground truth mesh in terms of the number of views selected across different $\alpha$ values.
As shown in Table~\ref{tab_reference_views}, the average discrepancy in the number of views between the meshes generated with different reference views and the ground truth, across all $\alpha$ values, is smallest when input with 3 reference views and remains similar with 5 reference views.

These two experiments confirm the advantage of conditioning diffusion model on multi-view images to generate more accurate 3D meshes, while also demonstrating that 3 reference views are sufficient for effective one-shot view planning

\begin{table}[!t]
\centering
\resizebox{0.65\columnwidth}{!}{%
\begin{tabular}{ccccc}
\toprule
\multirow{2}{*}{$\alpha$} & \multicolumn{4}{c}{EscherNet Reference Views} \\
                       & 1         & 2         & 3         & 5         \\
\midrule
1                      & 0.9       & \textbf{0.2}       & 0.4       & 1.1       \\
2                      & 1.5       & 1.2       & 1.1       & \textbf{1.0}       \\
3                      & 1.9       & 1.2       & 1.2       & \textbf{1.0}       \\
4                      & 3.0       & 1.3       & 1.3       & \textbf{1.1}      \\
5                      & 2.3       & 1.7       & 1.8       & \textbf{1.5}       \\
6                      & 2.9       & 1.8       & 1.8       & \textbf{1.6}       \\
7                      & 3.6       & 2.7       & \textbf{2.1}       & 2.2       \\
8                      & 4.8       & 3.1       & \textbf{2.9}       & 3.2       \\
9                      & 4.8       & 3.2       & 2.8       & \textbf{2.3}       \\
10                     & 4.4       & 3.6       & 3.0       & \textbf{2.9}       \\
11                     & 4.8       & 3.3       & \textbf{2.8}       & 3.9       \\
12                     & 5.1       & 4.9       & \textbf{3.3}       & 4.1       \\
13                     & 5.8       & 4.6       & \textbf{4.4}       & 4.6       \\
14                     & 6.7       & 5.6       & \textbf{5.3}       & 5.5       \\
15                     & 6.6       & 5.0       & 6.0       & \textbf{5.7}       \\
\midrule
Average                & 3.94      & 2.89      & \textbf{2.68}      & 2.78      \\
\bottomrule
\end{tabular}
}
\vspace{0.2cm}
\caption{
Evaluation of the impact of different number of reference views on the quality of the generated meshes.
Each value represents the absolute difference in the number of views selected by the proposed planner when using ground truth meshes versus EscherNet-generated meshes, averaged across 10 objects.
Lower discrepancy values indicate similar planning behavior compared to planning based on ground truth mesh, therefore indicating better quality of meshes generated by EscherNet. 
}
\label{tab_reference_views}
\end{table}

\subsection{Evaluation of View Planning Performance} 
\label{subsec:Evaluation}

In this section, we compare our method against both RGB-based one-shot and next-best-view planning baselines:
\begin{itemize}[leftmargin=*, noitemsep, topsep=0pt]
\item PRV~\citep{pan2024icra} employs a network to predict the number of views needed to achieve peak performance in NeRF training, using several reference views as input. 
It subsequently generates a fixed hemispherical view configuration based on the predicted number of views. 
For a fair comparison, we also use 3 reference views, consistent with our method.
\item Ensemble-NBV~\citep{lin2022rssworkshop} utilizes the variance of rendered RGB from NeRF ensembles to quantify uncertainty quantification to plan the NBV that maximizes the information gain.
The rendering resolution is set to $128 \times 96$, the number of ensembles is set to 3, and the training steps are set to 3000.
\item EscherNet-NBV employs EscherNet to generate a candidate view RGB image and calculates the mean squared error between this image and the same candidate view rendered by Instant-NGP trained on the same image collection.
The view with the largest difference is selected as the NBV. 
The image resolution is set to $256 \times 256$, consistent with EscherNet's configuration, and the training steps are set to 3000.
We implement this baseline to evaluate whether EscherNet could be effectively applied for NBV planning.
\end{itemize}

\subsubsection{Comparison to Baselines} 

Fig.~\ref{fig_baseline_ingp} reports the reconstruction quality and efficiency of different view planning methods using Instant-NGP.
When compared to NBV methods, our method achieves equal or better reconstruction quality while substantially reducing both movement cost and planning time.
The substantial reductions in on-board resources can be attributed to our global path planning and one-shot non-iterative approach, which eliminates the need for iterative map updates and uncertainty calculations. 
This underscores our main contribution to enabling object-specific one-shot view planning for RGB-based reconstruction by exploiting 3D diffusion model priors.
When compared to the one-shot PRV method, our approach reduces movement costs and offers greater flexibility in view planning, allowing for the prediction of fewer or more views to accommodate varying reconstruction quality requirements across different applications.
In summary, our method balances well between reconstruction quality and on-board resources.

\subsubsection{Generalization to Different Map Representations} 

\begin{figure*}[!t]
\centering
\includegraphics[width=1.0\textwidth]{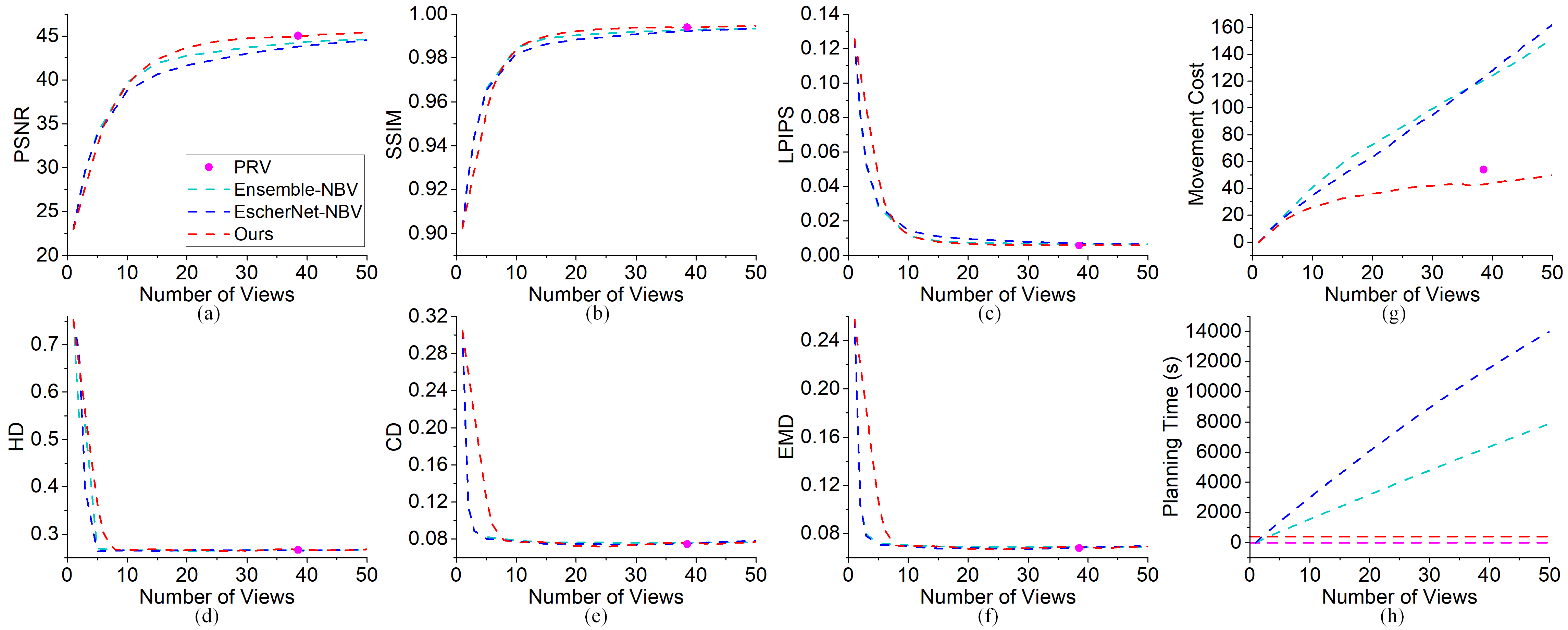}
\caption{
Evaluation of view planning performance using Instant-NGP.
For our planner, linear interpolation is then applied to draw the curves as the same in Fig.~\ref{fig_planner_ablation}.
For PRV, which predicts the same number of views based on the same reference views, its performance is represented by a single dot.
Each curve represents the average performance across 10 test objects with 2 different initial views.
As can be seen, our method achieves an overall higher or on-par reconstruction quality while substantially reducing the movement cost and requiring a reasonable planning time.
} 
\label{fig_baseline_ingp}
\end{figure*}

\begin{figure*}[!t]
\centering
\includegraphics[width=1.0\textwidth]{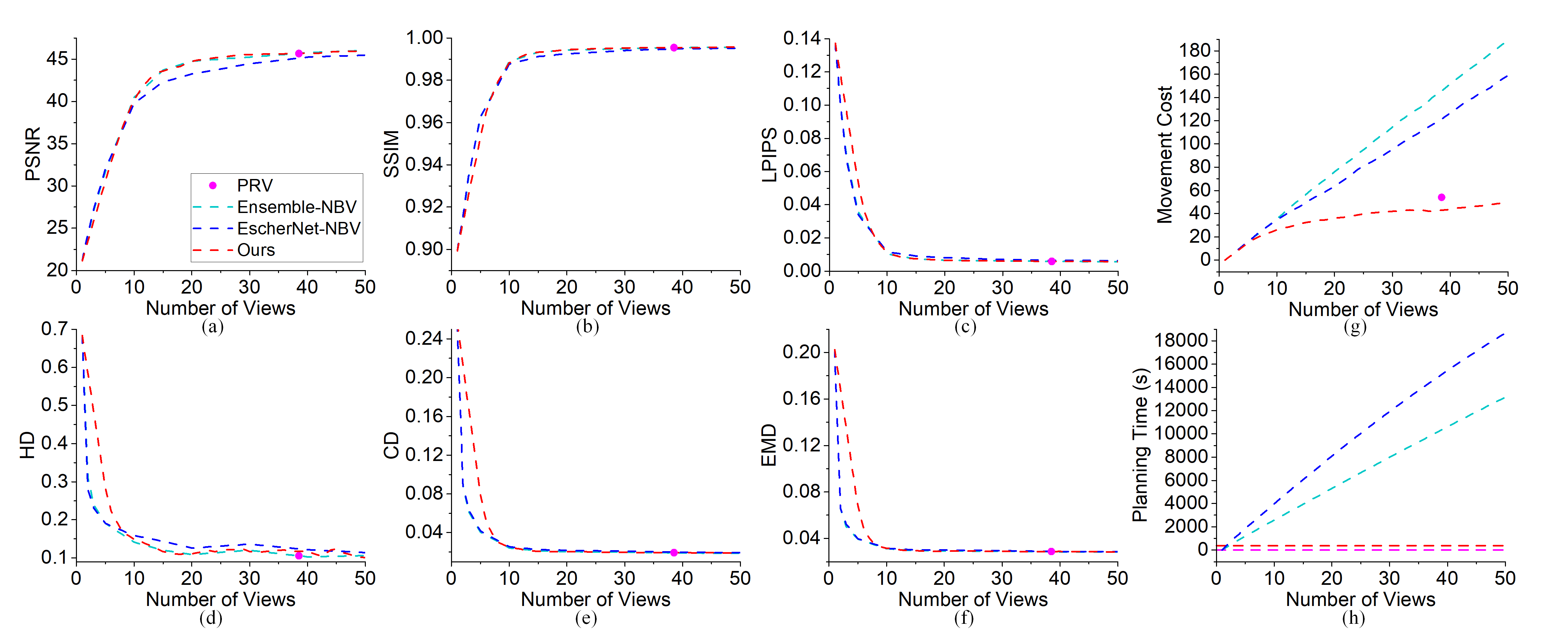}
\caption{
Evaluation of view planning performance using NeuS2. The experimental setup is consistent with that described in Fig.~\ref{fig_baseline_ingp}.
} 
\label{fig_baseline_neus2}
\end{figure*}

\begin{figure*}[!t]
\centering
\includegraphics[width=1.0\textwidth]{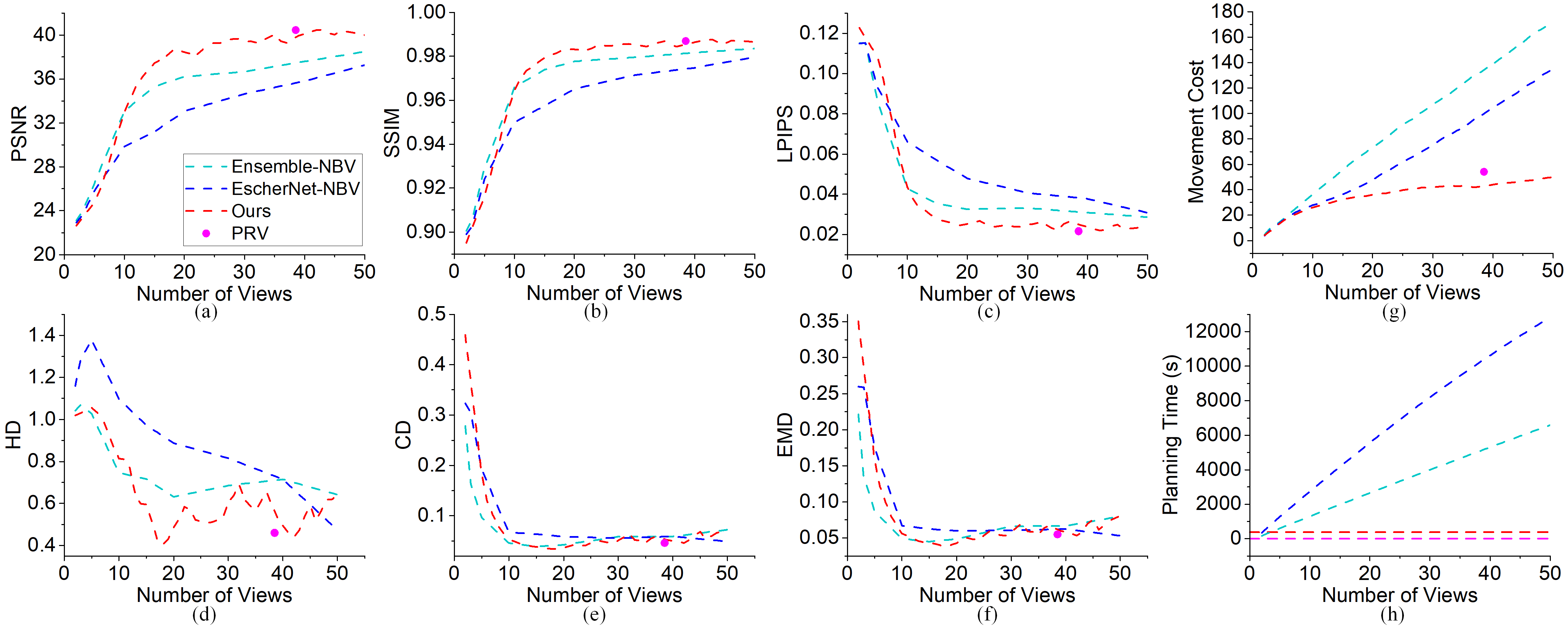}
\caption{
Evaluation of view planning performance using 2DGS. The experimental setup is consistent with that described in Fig.~\ref{fig_baseline_ingp}.
} 
\label{fig_baseline_2dgs}
\end{figure*}

Our system adopts Instant-NGP as the RGB-based reconstruction method after image collection.
However, we also show that our system can be generalized to different RGB-based reconstruction methods.
To this end, we evaluate the view planning performance on reconstruction using NeuS2~\citep{wang2023neus2} and 2DGS~\citep{huang20242d}, as shown in Fig.~\ref{fig_baseline_neus2} and Fig.~\ref{fig_baseline_2dgs}.
The results show that the performance of our one-shot view planning is consistent with that in Fig.~\ref{fig_baseline_ingp}.
This indicates the utilization of diffusion priors for out one-shot view planing is generalizable to different map representations, justifying broader application scenarios for our approach.

Although NeuS2 and 2DGS demonstrate better mesh geometry performance in terms of CD and EMD, their deployment in our real-world setup is hindered by unavoidable noise in recorded view poses, which disrupts the training process.
Unlike Instant-NGP, they do not support joint view pose optimization.
Instead, they rely on ColMap~\citep{fisher2021colmap} to compute view poses before training, which is unsuitable for our sparse view configuration as ColMap requires a dense view configuration.
to estimate poses accurately.
Therefore, we continue to use Instant-NGP in the following real-world experiments.

\subsection{Real-World Experiments}
\label{subsec:RealWorld}

We deploy our system in a real-world tabletop environment using a UR5 robotic arm equipped with a calibrated Intel RealSense D435 camera mounted on its end-effector, with only the RGB optical camera activated.
Different from our simulation experiments that assume a prefixed object-centric view space, we activate the localization module in our pipeline in real-world experiments to dynamically allocate the view space around the object.
This is necessary for our real-world setup, where both the object's position and scale are not known as a prior. 

\subsubsection{Study on Carving Views}
\label{subsebsec:CarvingViews}

\begin{table}[!t]
\centering
\resizebox{1.0\columnwidth}{!}{%
\begin{tabular}{c|cccc}
\hline
\multirow{2}{*}{Object Information} & \multicolumn{4}{c}{Number of Carving Views} \\
                                    &   1    & 2             & 3            & 4            \\ \hline
Center Displacement (cm)          &   8.92    & 6.59          & 0.48         & 0.15         \\ 
Size Change (cm)        &   15.16   & 12.00         & 0.57         & 0.41        \\
\hline
\end{tabular}%
}
\vspace{0.2cm}
\caption{Object localization results.
Each value is reported as the difference compared to the previous number of carving views.
For instance, the data for the third carving view is calculated as the difference relative to the second carving view. 
The initial object center is assumed to be at the center of the tabletop space, with an initial size of 10\,cm.
}
\label{tab_carving_number}
\end{table}

The accuracy of object localization is influenced by the number of carving views used. 
To determine the necessary number of carving views, we report the displacement of the object's center and the absolute change in its size after each carving step, as shown in Table~\ref{tab_carving_number}.
The results indicate large differences in the center displacement and size changes when coming to the first and second carving views.
However, these differences decrease substantially when comparing two views to three views and three views to four views, with changes smaller than 1\,cm.
This is reasonable, as two view cones are likely to carve out the majority of the surrounding area of the object.
Based on this observation, we select two carving views as our configuration, as it provides sufficient initial positioning accuracy while minimizing resource consumption.

\subsubsection{Dynamic View Space Generation}

\begin{figure*}[!t]
\centering
\includegraphics[width=0.75\textwidth]{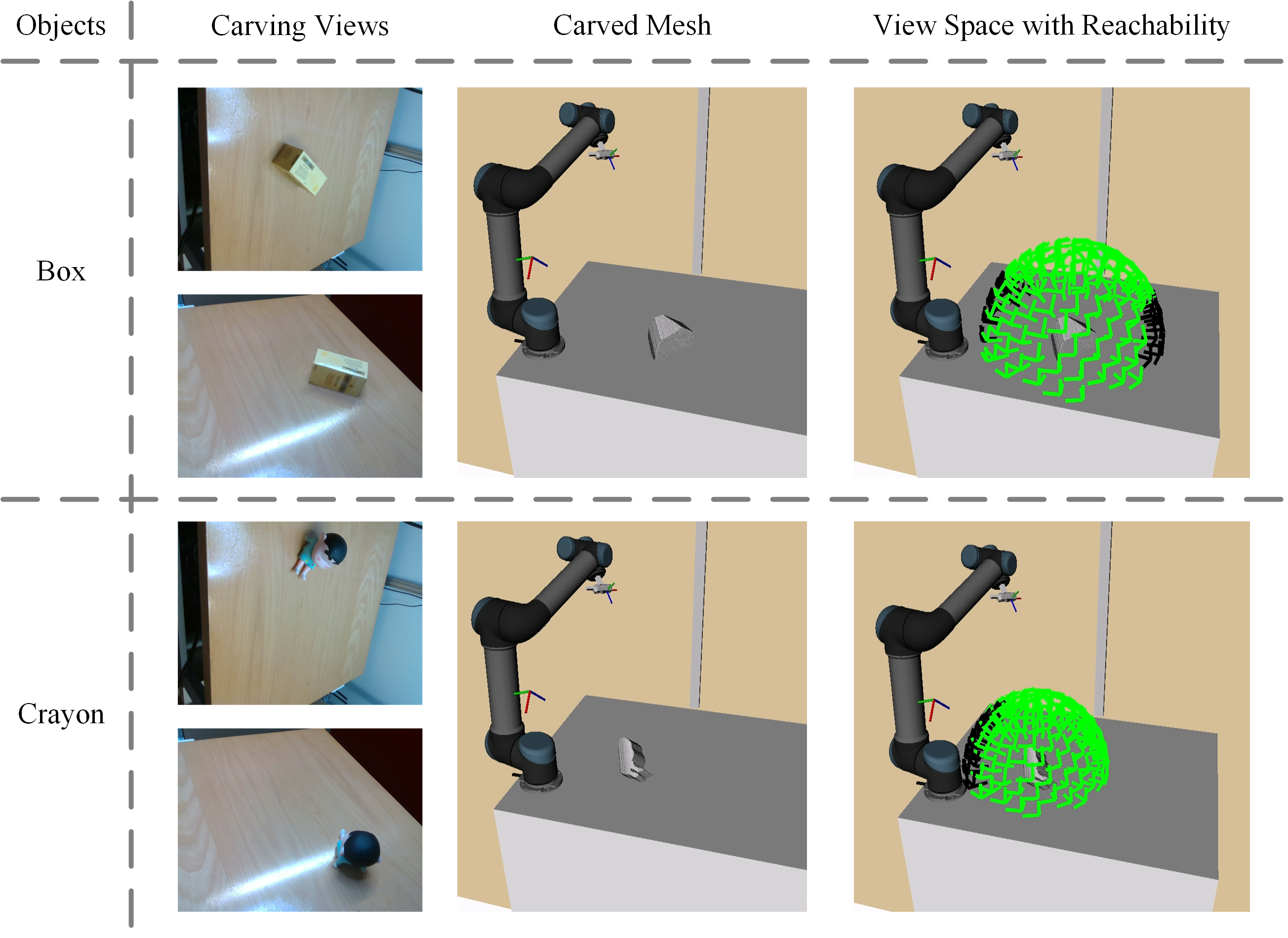}
\caption{
Dynamic generation of the view space based on varying object positions and sizes.
As can be seen, (1) the center of the view space adjusts dynamically based on the object's position; (2) the smaller cartoon Crayon object results in a smaller radius of the view space with a higher number of reachable views (green) compared to the larger Box object.
} 
\label{fig_carving_motion}
\end{figure*}

The rough position and size of the object are obtained from our localization module, accommodating variations in object position and size.
To demonstrate this, we present qualitative results of dynamic view space generation.
We test three objects: the Flower object shown in Fig.~\ref{fig_pipeline}, and two additional objects illustrated in Fig.~\ref{fig_carving_motion}. 
The integration of the localization module effectively enables dynamic placement of the object-centric view space.
This adaptability allows the system to accommodate objects of varying sizes and arbitrary positions without prior knowledge of their location or dimensions.
Additionally, our method discards unreachable views identified by MoveIt after the discretization of the view space. 
This dynamic adjustment ensures better alignment with physical constraints, enhancing the system's flexibility and practicality in real-world applications.

\subsubsection{Ground Truth Approximation}

In the real world, unlike in simulation where the ground truth model is available, the object's 3D model is inaccessible.
To address this, we utilize all images from the entire planning view space to train a NeRF, generating a 3D mesh as an approximation of the ground truth model for evaluation.
This approach is reasonable as it sets an upper bound on the achievable performance of view planning, which aims to select an optimized subset of the discretized view space.
Additionally, the novel test view images are collected by the robot, with their poses refined through joint pose optimization across the entire planning view space, ensuring a reliable quantitative comparison.

\subsubsection{Comparison to Baselines}

\begin{figure*}[!t]
\centering
\includegraphics[width=1.0\textwidth]{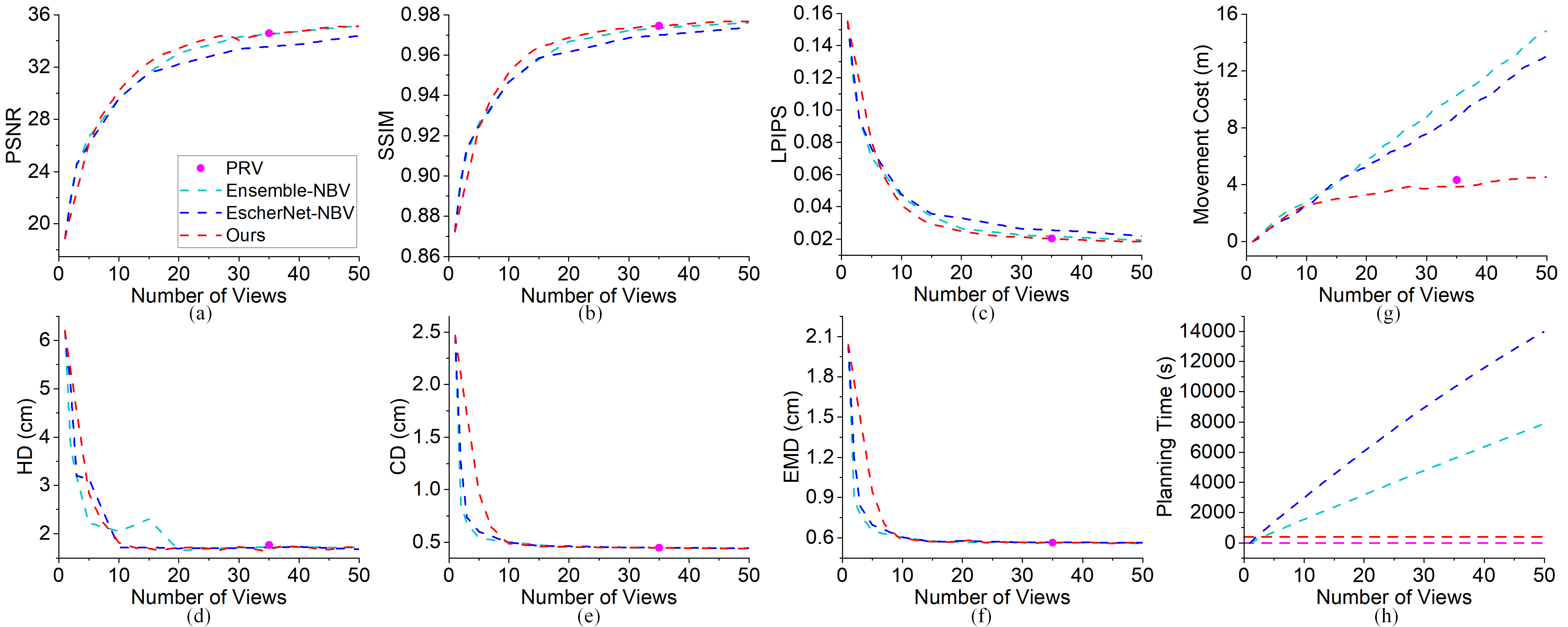}
\caption{
Evaluation of view planning performance in the real world. Each curve represents the average performance across 3 real-world objects.
} 
\label{fig_real_baseline}
\end{figure*}

To validate our findings from simulation, we compare our method against baselines in real-world experiments.
The experimental comparisons are presented in Fig.~\ref{fig_real_baseline}.
The results demonstrate that the performance of our one-shot view planning is consistent with the findings in simulation, confirming its generalizability to real-world environments.
Notably, we observe a decrease in the PSNR upper limit from approximately 45 in simulation to around 35 in real-world settings.
This drop can likely be attributed to real-world factors such as lighting variations, noise in background removal, and aliasing effects.
Nonetheless, achieving a PSNR above 30 is sufficient for practical applications, highlighting the robustness of our approach under real-world conditions.

\subsection{Analysis of Multi-View Constraints}
\label{subsec:Multi-View Constraints}

\begin{table*}[!t]
\centering
\resizebox{0.8\textwidth}{!}{%
\begin{tabular}{cccccccccc}
\toprule
Alpha & PSNR  & SSIM   & LPIPS  & HD     & CD     & EMD    & \makecell{Number\\of Views} &  \makecell{Movement\\Cost} &  \makecell{Planning\\Time\,(s)} \\
\midrule
1     & 35.14 & 0.9698 & 0.0239 & 0.2670 & 0.0808 & 0.0710 & 6.2             & 19.09         & 389.7         \\
2     & 39.91 & 0.9853 & 0.0110 & 0.2667 & 0.0761 & 0.0694 & 10.5            & 26.80         & 389.8         \\
3     & 41.56 & 0.9886 & 0.0086 & 0.2665 & 0.0777 & 0.0701 & 13.1            & 30.04         & 390.1         \\
4     & 42.38 & 0.9897 & 0.0080 & 0.2695 & 0.0764 & 0.0691 & 15.3            & 32.59         & 390.4         \\
5     & 43.24 & 0.9917 & 0.0069 & 0.2656 & 0.0752 & 0.0685 & 18.4            & 35.39         & 390.8         \\
\textbf{6}     & 43.79 & 0.9924 & 0.0066 & 0.2659 & 0.0723 & 0.0674 & 20.9            & 36.18         & 390.8         \\
7     & 44.21 & 0.9930 & 0.0063 & 0.2641 & 0.0726 & 0.0674 & 23.9            & 38.93         & 391.0         \\
8     & 44.48 & 0.9934 & 0.0061 & 0.2650 & 0.0719 & 0.0670 & 26.4            & 40.57         & 391.2         \\
9     & 44.67 & 0.9937 & 0.0061 & 0.2668 & 0.0736 & 0.0677 & 29.2            & 41.53         & 390.8         \\
\textbf{10}    & 44.83 & 0.9940 & 0.0060 & 0.2644 & 0.0730 & 0.0675 & 32.0            & 42.04         & 390.7         \\
11    & 44.86 & 0.9939 & 0.0060 & 0.2656 & 0.0742 & 0.0683 & 33.7            & 42.43         & 390.4         \\
12    & 44.98 & 0.9942 & 0.0058 & 0.2679 & 0.0758 & 0.0686 & 36.6            & 43.25         & 390.3         \\
13    & 44.96 & 0.9939 & 0.0062 & 0.2676 & 0.0766 & 0.0695 & 38.7            & 43.89         & 390.1         \\
14    & 45.15 & 0.9942 & 0.0060 & 0.2656 & 0.0753 & 0.0684 & 42.0            & 45.03         & 390.1         \\
15    & 45.22 & 0.9942 & 0.0061 & 0.2673 & 0.0759 & 0.0690 & 45.8            & 47.47         & 390.1         \\
\bottomrule
\end{tabular}
}
\vspace{0.2cm}
\caption{
Evaluation of view planning performance under different $\alpha$ values for our one-shot view planning method in simulation.
Each value represents the average across 10 test objects, each tested with 2 different initial views.
The results reveal the following: (1) onboard resource usage (the number of views and movement cost) increases as $\alpha$ grows; (2) mesh quality metrics converge at $\alpha = 6$, indicating sufficient reconstruction quality with fewer resources; and (3) image quality peaks at $\alpha = 10$, with diminishing improvements beyond this point (\eg, the PSNR improvement from $\alpha = 10$ to $\alpha = 11$ is less than 0.1).
}
\label{tab_alpha}
\end{table*}

In this section, we examine the impact of multi-view constraints on reconstruction performance and provide practical recommendations for deploying our system.
In real applications, the parameter $\alpha$ has to be set in advance to achieve the desired view planning outcomes.
We first evaluate the $\alpha$ in simulation in Table~\ref{tab_alpha}.
The results indicate that with increasing $\alpha$ values, our
optimizer outputs on average more views for covering the mesh surfaces.
For $\alpha = 6$, our method achieves strong performance across image metrics, including PSNR, SSIM, and LPIPS, as well as effectively converged geometric metrics, such as HD, CD, and EMD.
With an average of 20.9 views and a low movement cost of 36.18, $\alpha = 6$ demonstrates feasibility for sparse 3D object reconstruction.
When $\alpha = 10$, while PSNR improves slightly from 43.79 to 44.83, the mesh-based metrics show negligible gains compared to $\alpha = 6$.
Based on these findings, we recommend setting $\alpha = 6$ for more efficient object reconstruction, as it strikes a good balance between performance and resource consumption.
However, for applications requiring more realistic visuals in novel view rendering, $\alpha = 10$ may be preferred despite the marginal improvement in mesh reconstruction metrics.

\begin{figure*}[!t]
\centering
\includegraphics[width=1.0\textwidth]{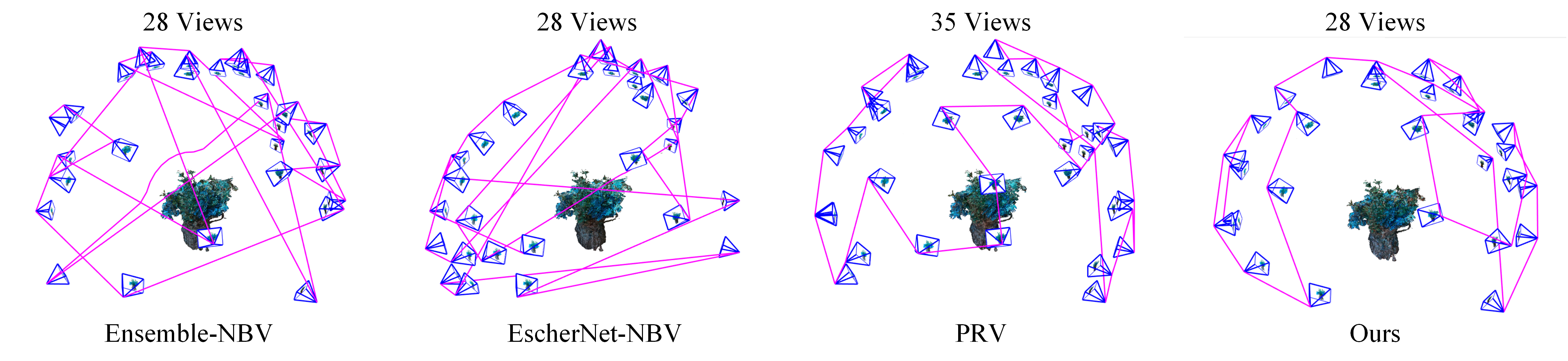}
\caption{
Illustration of planned views, paths, and reconstructed meshes of the Flower object.
Our method uses $\alpha = 6$ in this example.
For a fair comparison, the NBV methods are shown using the same number of views as those selected by our method.
} 
\label{fig_compare_flower}
\end{figure*}

\begin{table*}[!t]
\centering
\resizebox{1.0\textwidth}{!}{%
\begin{tabular}{cccccccccc}
\toprule
Method       & Number of Views                & PSNR  & SSIM   & LPIPS  & HD (cm) & CD (cm) & EMD (cm) & Movement Cost (m) & Planning Time (s) \\
\midrule
Ensemble-NBV & \multirow{3}{*}{28 ($\alpha=6$)}  & 30.82 & 0.9472 & 0.0304 & 2.413   & 0.475   & 0.559    & 7.25           & 4647.8         \\
EscherNet-NBV &    & 30.75 & 0.9464 & 0.0305 & 2.389 & 0.476 & 0.557 & 7.62 & 8753.8  \\
Ours          &    & 31.27 & 0.9497 & 0.0287 & 2.384 & 0.476 & 0.569 & \textbf{3.82} & 429.3  \\
\midrule
PRV           & 35 & 31.68 & 0.9532 & 0.0264 & 2.383 & 0.471 & 0.562 & 4.32 & \textbf{0.033}    \\
\midrule
Ensemble-NBV & \multirow{3}{*}{43 ($\alpha=10$)} & 31.76 & 0.9545 & 0.0259 & 2.396   & 0.466   & \textbf{0.552}    & 11.67           & 7090.7          \\
EscherNet-NBV &    & 31.74 & 0.9541 & 0.0257 & 2.365 & 0.464 & 0.556 & 12.5515 & 12931.7 \\
Ours          &    & \textbf{32.00} & \textbf{0.9560} & \textbf{0.0250} & \textbf{2.346} & \textbf{0.464} & 0.562 & 4.50 & 427.5 \\
\bottomrule
\end{tabular}%
}
\vspace{0.2cm}
\caption{Evaluation of the view planning performance of the Flower object (as shown in Fig.~\ref{fig_pipeline}) under $\alpha = 6$ and $\alpha = 10$.
For a fair comparison, the NBV methods are evaluated using the same number of views as those selected by our method.
As can be seen, our method selects more views to accommodate the complexity of the Flower object and achieves the desired performance at both $\alpha = 6$ and $\alpha = 10$, further validating the suitability of our recommended $\alpha$ values.
}
\label{tab_flower}
\end{table*}

Next, we validate our recommendation of using $\alpha = 6$ and $\alpha = 10$ for one-shot view planning with the representative real-world Flower object, chosen for its high geometric and textural complexity. 
We compare our method against baselines at these two $\alpha$ values shown in Table~\ref{tab_flower} and visualize $\alpha=6$ in Fig.~\ref{fig_compare_flower}.
Compared to NBV methods, our approach outperforms them at both $\alpha = 6$ and $\alpha = 10$, achieving better image and mesh quality while substantially reducing planning time and movement cost.
When compared to PRV, our method demonstrates greater flexibility.
At $\alpha = 6$, it achieves a more sparse reconstruction with lower movement cost and sufficient reconstruction quality.
Although the PRV network predicts 43 views for this example, some viewpoints in its fixed view space are unreachable by the robot.
Consequently, only 35 views are moved for image collection, falling short of achieving comparable performance to our method at $\alpha = 10$.
In summary, these results further validate our method's flexibility in real-world scenarios and its ability to achieve a good balance between reconstruction quality and efficiency with recommended $\alpha$ values.

\section{Discussion} \label{sec:Discussion}

\begin{table}[!t]
\centering
\resizebox{\columnwidth}{!}{%
\begin{tabular}{cccccc}
\toprule
\makecell{Planner\\Components} & \makecell{Diffusion\\Generation} & \makecell{Surface\\Sampling} & \makecell{Ray\\Casting} & \makecell{Complexity\\Computation} & \makecell{Covering\\Optimization} \\
\midrule
Runtime\,(s)      & 308.3                & 12.1             & 65.2        & 2.2                    & 2.5              \\
\bottomrule
\end{tabular}%
}
\vspace{0.2cm}
\caption{Runtime of each component of our one-shot view planning method. Since the planning time does not vary much across different $\alpha$ values as shown in Table~\ref{tab_alpha}, we report the averaged time for each component across all $\alpha$ values.
}
\label{tab_time_element}
\end{table}

Although our one-shot view planning method leveraging 3D diffusion models achieves promising results in both simulation and real-world scenarios, there remains room for further improvement.
One limitation of our approach lies in the planning time.
While our method substantially reduces planning time compared to NBV methods, it requires more time than PRV, which leverages a lightweight network to predict the required number of views.
To further examine the planning time, we break it down into several components, as presented in Table~\ref{tab_time_element}.
The primary contributor to planning time is EscherNet. 
Since EscherNet employs a two-stage process—generating images and subsequently training NeuS2 for 3D generation—the process is inherently slow.
One potential improvement is to design a framework that incorporates EscherNet’s view pose encoding but operates more like One-2-3-45++, which performs 3D generation directly in 3D space and requires only about 45 seconds for the generation.
Another time-consuming component is ray casting using OctoMap.
This process could be optimized in the future with GPU acceleration, as it is highly parallelizable.
On the other hand, extending this work to a multi-robot active reconstruction system, as explored in~\citep{dhami2023map}, holds great promise and could mitigate the impact of the minute-level inference time.
In such a system, multiple robots could begin by capturing several reference images for the diffusion models and then wait for the computation server to provide the view planning results.
This parallelized approach would allow the robots to utilize their idle time effectively, potentially reducing overall mission time and improving system efficiency.

\section{Conclusion} \label{sec:Conclusion}

In this paper, we present an active RGB-based reconstruction system powered by a novel one-shot view planning method that leverages priors from 3D diffusion models using a few reference images.
The adoption of a multi-image-to-3D diffusion model addresses hallucination issues associated with a single-image input.
The generated 3D mesh serves as a proxy for the inaccessible ground truth 3D model, forming the basis for one-shot view planning.
We introduce a customized set covering optimization problem tailored for RGB-based reconstruction, aiming to compute an object-specific view configuration that densely covers the generated mesh based on its local geometric and textural complexity.
A globally shortest path is computed on this view configuration, minimizing travel distance for data collection.
Our simulation experiments demonstrate that leveraging priors from 3D diffusion models produces more informative views, while the one-shot paradigm achieves an excellent balance between onboard resource usage and reconstruction quality.
Testing with different RGB-based reconstruction methods, such as NeuS2 and 2DGS, highlights the broader applicability of the proposed method beyond NeRF.
In real-world scenarios, we integrate voxel carving to localize objects and test view planning under the physical constraints of the robot.
These underscore the practicality and flexibility of our system, showcasing its potential for diverse applications.

\begin{dci}
The authors declared no potential conflicts of interest with respect to the research, authorship, and/or publication of this article.
\end{dci}

\begin{funding}
This work has partially been funded by the Deutsche Forschungsgemeinschaft (DFG, German Research Foundation) under grant 459376902 – AID4Crops, under Germany’s Excellence Strategy, EXC-2070 – 390732324 – PhenoRob, by the EC, grant No. 964854 RePAIR H2020-FETOPEN-2018-2020 and by the BMBF within the Robotics Institute Germany, grant No. 16ME0999.
\end{funding}

\appendix
\section{Uniform-Aimed One-Shot View Planner with Object Complexity}
\label{apd:planner_uniform}

In this appendix, we present the integer linear programming formulation and the search algorithm for the uniform-aimed covering planner, referred to as Uni+Com.
This additional one-shot view planner is introduced to evaluate whether spatially uniform coverage is sufficient for object-centric view planning.
The goal of Uni+Com is to achieve the most spatially uniform distribution of views while considering object complexity:

\vspace{-0.2cm}
\begin{equation}
\label{equ:ILP}
\begin{aligned}
\max: \, & \sum_{v_i\in \mathcal{V}} \sum_{v_j \in \mathcal{V} \wedge j \neq i} d^i_j\, x_{v_i}\, x_{v_j}  \, ,\\
\mathrm{s.t.}: \,
& (a) \, x_{v} \in \{0,1\}   && \forall v \in \mathcal{V} \\
& (b) \sum_{v\in \mathcal{V}} x_v = N, && \forall v \in \mathcal{V}
\end{aligned}
\end{equation}
where the objective function $\sum_{v_i\in \mathcal{V}} \sum_{v_j \in \mathcal{V} \wedge j \neq i} d^i_j\, x_{v_i}\, x_{v_j}$ is to maximize the total distance between any two selected views, where $d^i_j$ is the Euclidean distance between views $v_i$ and $v_j$; while subject to two constraints: (a) $x_{v}$ is a binary variable representing whether a view $v$ is included in the set of selected views or not; and (b) the number $\sum_{v\in \mathcal{V}} x_v$ of selected views equals to an integer value $N$.

To plan a set of views using this formulation, we incrementally increase $N$ from $1$ to $|\mathcal{V}|$ until the multi-view constraints based on object complexity are satisfied.
However, the constraints $\sum_{v \in \mathcal{V}} I(p, v)\, x_{v} \geq \left\lceil \alpha C(p) \right\rceil,\, \forall p \in \mathcal{P}_{\mathit{surf}}$ often conflict with the objective function.
In other words, directly incorporating the multi-view constraints as hard constraints into the optimization problem would result in a similar performance to our SetRaw+Uni+Com planner.
The Uni+Com planner, however, aims to prioritize a spatially uniform distribution of views.
To achieve this, we relax the constraints by making them soft: instead of enforcing all constraints strictly, we ensure that at least 95\% of the surface points satisfy the complexity coverage requirements.
This adjustment allows the planner to maintain its focus on spatial uniformity while still addressing object complexity.

Additionally, the objective function for achieving spatially uniform distribution is effective only for symmetric spheres. 
To address the subset of the sphere representing the reachable viewpoints, we first solve the problem assuming a full sphere and then filter out the unreachable viewpoints during post-processing.
This ensures that the planner works effectively within the hemispherical view space or robot's physical constraints while still adhering to its goal of spatial uniformity.

\bibliographystyle{SageH}
\bibliography{ijrr2024}

\end{document}